%% file: main.tex
\documentclass[times,review,10pt]{elsarticle}

\usepackage{amssymb}
\usepackage{amsmath}

\usepackage{xcolor}
\usepackage{multirow}
\usepackage{booktabs}
\usepackage{ulem}
\usepackage{graphicx}
\usepackage{float}
\usepackage{subfig}
\usepackage{bbding}
\usepackage{rotating}

\journal{arXiv}

\begin{document}

\begin{frontmatter}



\title{RhythmFormer: Extracting Patterned rPPG Signals based on Periodic Sparse Attention}

\author{Bochao Zou$^a$, Zizheng Guo$^a$, Jiansheng Chen$^a$, Junbao Zhuo$^a$, Weiran Huang$^b$, Huimin Ma$^{a,}$\corref{cor1}}

\cortext[cor1]{Corresponding author.}

\affiliation{organization={SCCE, University of Science and Technology Beijing},
            postcode={100083}, 
            state={Beijing},
            country={China}}
\affiliation{organization={Qing Yuan Research Institute, SEIEE, Shanghai Jiao Tong University},
            postcode={200240}, 
            state={Shanghai},
            country={China}}
\begin{abstract}
  Remote photoplethysmography (rPPG) is a non-contact method for detecting physiological signals based on facial videos, holding high potential in various applications. Due to the periodicity nature of rPPG signals, the long-range dependency capturing capacity of the transformer was assumed to be advantageous for such signals. However, existing methods have not conclusively demonstrated the superior performance of transformers over traditional convolutional neural networks. This may be attributed to the quadratic scaling exhibited by transformer with sequence length, resulting in coarse-grained feature extraction, which in turn affects robustness and generalization. To address that, this paper proposes a periodic sparse attention mechanism based on temporal attention sparsity induced by periodicity. A pre-attention stage is introduced before the conventional attention mechanism. This stage learns periodic patterns to filter out a large number of irrelevant attention computations, thus enabling fine-grained feature extraction. Moreover, to address the issue of fine-grained features being more susceptible to noise interference, a fusion stem is proposed to effectively guide self-attention towards rPPG features. It can be easily integrated into existing methods to enhance their performance. Extensive experiments show that the proposed method achieves state-of-the-art performance in both intra-dataset and cross-dataset evaluations. The codes are available at https://github.com/zizheng-guo/RhythmFormer.
\end{abstract}

\begin{keyword}
Remote physiological measurement \sep Periodic sparse attention


\end{keyword}

\end{frontmatter}



\input{sec/1_intro}
\input{sec/2_relatedwork}
\input{sec/3_methodology}
\input{sec/4_experiment}
\input{sec/5_conclusion}

\section*{Acknowledgments}
This work was supported in part by the National Natural Science Foundation of China (62206015, 62227801, 62376024, 62402033), the National Science and Technology Major Project (2022ZD0117901), the Fundamental Research Funds for the Central Universities (FRF-TP-22-043A1), and the Young Scientist Program of The National New Energy Vehicle Technology Innovation Center (Xiamen Branch).

\bibliographystyle{elsarticle-num} 
\bibliography{main}

\end{document}

%% file: sec/1_intro.tex
\section{Introduction}
\label{sec:intro}

Blood Volume Pulse (BVP) is an important physiological signal, further allowing for the extraction of key vital signs such as heart rate (HR) and heart rate variability (HRV). Photoplethysmography (PPG) is a non-invasive monitoring method that utilizes optical means to measure changes in blood volume within living tissues. The physiological mechanism underlying BVP extraction from PPG is rooted in variations in light absorption and scattering due to changes in blood volume during cardiac contraction and relaxation in subcutaneous blood vessels. These variations result in periodic color signal alterations on imaging sensors, imperceptible to the human eye~\cite{GREEN,deepphys}. Traditionally, BVP extraction required the use of contact sensors~\cite{carrera2019pr,liu2022pr}, introducing inconveniences and limitations. In recent years, non-contact methods for extracting BVP information, specifically remote photoplethysmography (rPPG), have garnered increasing attention~\cite{pyvhr,review1}. 
These methods possess significant research value not only within the field of healthcare but also hold broad application prospects in a multitude of fields, including affective computing, deepfake detection, and anti-spoofing, among others~\cite{apptac23}.

\begin{figure}[t]
  \centering  
  \setlength{\abovecaptionskip}{0.1cm}
  \includegraphics[width=0.96\linewidth]{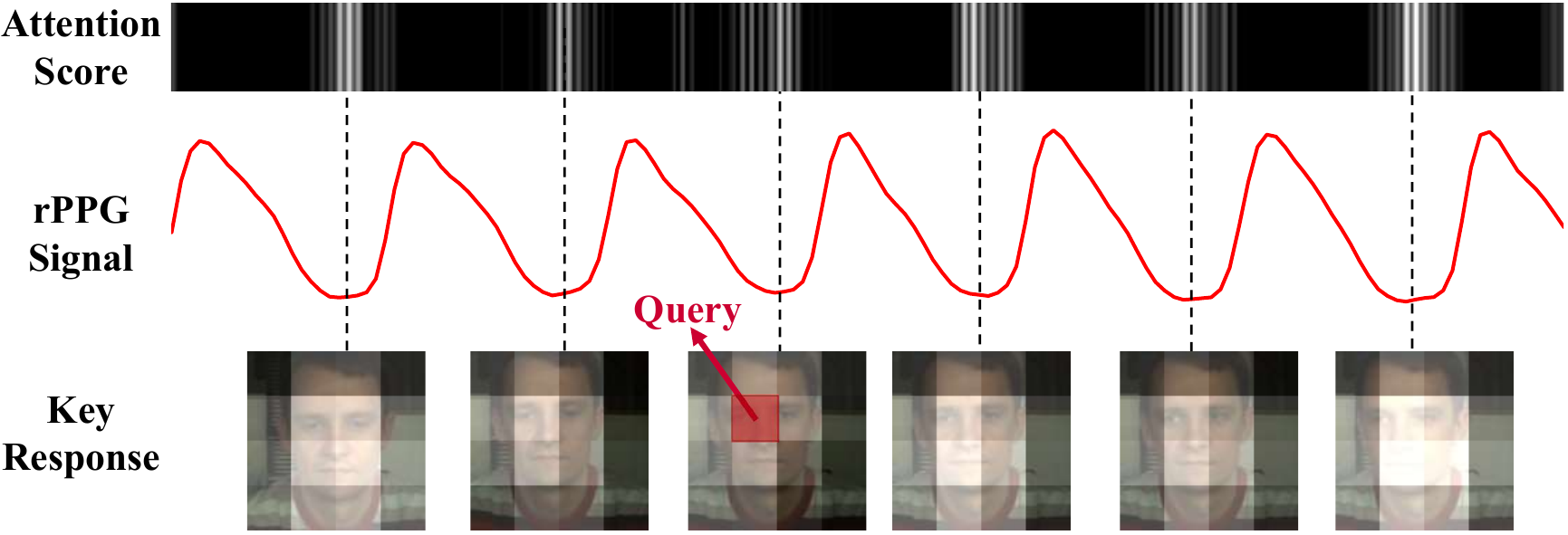}  
  \caption{Sparsity in attention due to periodicity. Given the red block as a Query, the global spatio-temporal correlations are computed. Brighter regions in the attention score (top row) and key response (bottom row) in the frame indicate stronger attention. From the figure, it is evident that the brightness of the key response in the frame is predominantly concentrated in the facial area, while the brightness in the attention score is mainly focused on the troughs (regions with the same phase as the Query). The attention scores exhibit a notable sparsity.}
  \label{fig:1}
\end{figure}

Early research on rPPG primarily relied on signal processing-based methods to analyze facial videos for extracting periodic signals~\cite{GREEN, POS, casado2023face2ppg}.
However, due to the weak nature of rPPG signals, which have low amplitude and are subject to strong noise and artifacts from environmental factors such as lighting and motion, achieving high accuracy solely through signal processing in complex environments is challenging.
With the advancement of deep learning, there is a growing trend towards data-driven methods. In recent years, significant attention has been directed towards the transformer within the rPPG field. Compared with convolutional neural networks (CNNs), transformers excel in capturing long-range dependency. This enables them to model BVP features by leveraging intra-frame (facial region) and inter-frame (similar temporal phase) correlations, making them highly advantageous for quasi-periodic rPPG signals.
However, existing research has not convincingly demonstrated that transformers outperform traditional CNNs. This may stem from their merely passive exploitation of the long-range dependency advantages for periodicity, which is achieved through global attention or static sparse attention. Nevertheless, the quadratic scaling exhibited by transformers with sequence length~\cite{vivit}, further compounded by the high dimensionality of video, restricted these approaches to transforming inputs into coarse-grained tokens.
Due to artifacts and head movement, these coarse tokens inevitably encompass both regions rich in rPPG information and regions with significant noise~\cite{physformer}. The introduction of noise through coarse-grained extraction, combined with the inherent weak nature of rPPG signals, results in a low signal-to-noise ratio, significantly impacting robustness and generalization.

\begin{figure*}[t]
  \centering  
  \includegraphics[width=1.00\linewidth]{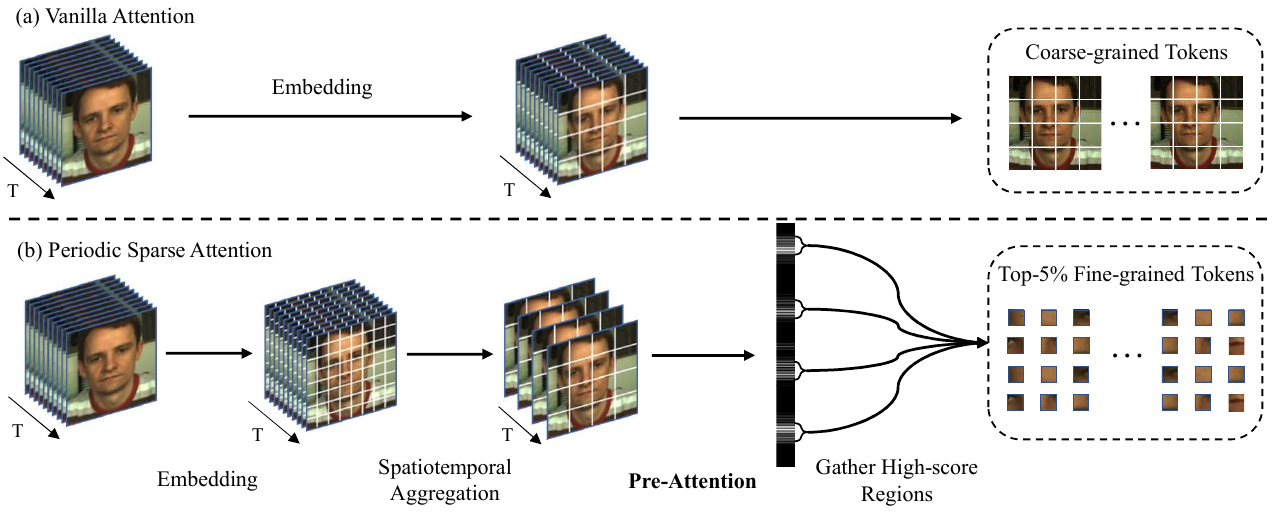}  
  \caption{The proposed periodic sparse attention mechanism utilizes the periodic patterns learned through pre-attention to gather high-score regions (facial regions with similar temporal phases), which filters out a substantial number of irrelevant attention computations. This allows for finer-grained token extraction across dimensions H, W, and T compared with vanilla attention.}
  \label{fig:12}
\end{figure*}

To address this challenge, we focus on the phenomenon of notable sparsity in temporal attention due to periodicity, as shown in Figure~\ref{fig:1}. Inspired by findings in cognitive psychology related to pre-attentive vision, which refers to the processing of visual stimuli before they are selected by attention for more complete analysis~\cite{wolfe2019preattentive,wolfe2021guided}, we propose a periodic sparse attention mechanism. This mechanism explicitly leverages the core periodic nature of rPPG, integrating it into the attention design. As illustrated in Figure~\ref{fig:12}, a pre-attention stage is introduced before the conventional attention mechanism.
The pre-attention stage learns potential periodic patterns in the rPPG signals and involves temporally aggregating regions with high phase similarity and spatially aggregating areas rich in rPPG signals, effectively filtering out numerous irrelevant attention computations. This allows subsequent attention to focus on fine-grained feature extraction.

Moreover, the inherent weak nature of rPPG signals and the susceptibility of fine-grained features to noise interference collectively lead to noticeable self-attention biases. To address this challenge, we devise a fusion stem. This design allows frame-level representations to capture BVP wave variations, effectively guiding the self-attention mechanism.

The main contributions are as follows:

$\bullet$ Leveraging the sparsity of temporal attention stemming from periodicity, we design a periodic sparse attention mechanism, which filters out redundant computations through pre-attention, enabling fine-grained attention.

$\bullet$ We propose a plug-and-play fusion stem module, enabling frame-level representation awareness of BVP wave variations. This module can be seamlessly integrated into other methods, significantly improving the performance of existing approaches.

$\bullet$ We propose RhythmFormer. Extensive experiments conducted on both intra-dataset and cross-dataset scenarios demonstrate that the proposed method outperforms previous state-of-the-art approaches.


%% file: sec/2_relatedwork.tex
\section{Related Work}
\label{sec:relatedwork}

\subsection{Remote Physiological Estimation}

Early research on rPPG predominantly relied on traditional signal processing methods to analyze facial videos for extracting quasi-periodic signals \cite{GREEN, CHROM, POS, casado2023face2ppg}. 
With the advancement of deep learning, an increasing number of data-driven approaches have emerged, showcasing a trend in encoder transitioning from 2D CNNs \cite{HR-CNN, rhythmNet, deepphys, TSCAN, zhang2025} to 3D CNNs \cite{physnet,3DCDC,contrastphys+,yue2023}  and further to transformer \cite{physformer, efficientphys,liu2024spikingphysformer}. For video-based time-series tasks such as rPPG, 3D CNNs offer a significant advantage over 2D CNNs in the extraction of temporal information. Compared to 3D CNNs, transformers excel in capturing long-range spatial and temporal correlations, which enables them to model BVP signals by leveraging intra-frame (skin region) and inter-frame (similar temporal phase) correlations.
However, many of the most recent self-supervised methods or applications still employ CNN architectures as encoders \cite{contrastphys+,yue2023,zhang2025}. 
For example, Greip~\cite{zhang2025} uses ResNet as the encoder through the integration of explicit and implicit prior knowledge to improve the generalization performance of the model. This may stem from the fact that transformers have not shown a significant performance advantage over CNNs, while CNNs are more user-friendly. The theoretical advantages of transformers have not been fully realized, potentially due to the constraints imposed by quadratic complexity in previous transformer-based approaches. These constraints have led to a coarse-grained extraction of features, thereby not fully leveraging the transformers' capacity for capturing dependencies.

Faced with the challenge of quadratic complexity, the proposed RhythmFormer not only passively leverages periodic advantages for long-range inter-frame attention, as in previous work, but also proactively exploits periodicity at a deeper level. It introduces a pre-attention stage that filters out a substantial amount of irrelevant attention computations through the learning of periodic patterns, thereby achieving more fine-grained attention at a lower computational cost.

\subsection{Sparse Attention Mechanism}

The conventional attention mechanism poses a significant computational burden and demands substantial memory resources as it involves pair-wise token interaction across all spatial-temporal locations \cite{transformer,vivit}. As an effective solution, sparse attention has gained recent popularity \cite{swintransformer,mvit,biformer}. 
Previous rPPG works \cite{physformer,efficientphys,liu2024spikingphysformer} primarily employed coarse-grained global attention without incorporating sparse attention mechanisms. This might be attributed that rPPG information is present in the skin regions across all temporal periods, making it difficult to achieve sparse attention through hand-crafted static attention mechanisms.

In contrast to global attention or static sparse attention in previous work, we embed the quasi-periodic nature of rPPG into the sparse attention mechanism. Based on the sparsity of temporal attention stemming from periodicity, we design a periodic sparse attention mechanism tailored to periodic tasks. This mechanism leverages periodicity to filter out a significant amount of redundant computations, enabling fine-grained sparse attention.

%% file: sec/3_methodology.tex
\label{sec/3_methodology}

\begin{figure*}[htbp] 
  \centering  
  \setlength{\abovecaptionskip}{0.25cm}
  \includegraphics[width=0.99\linewidth]{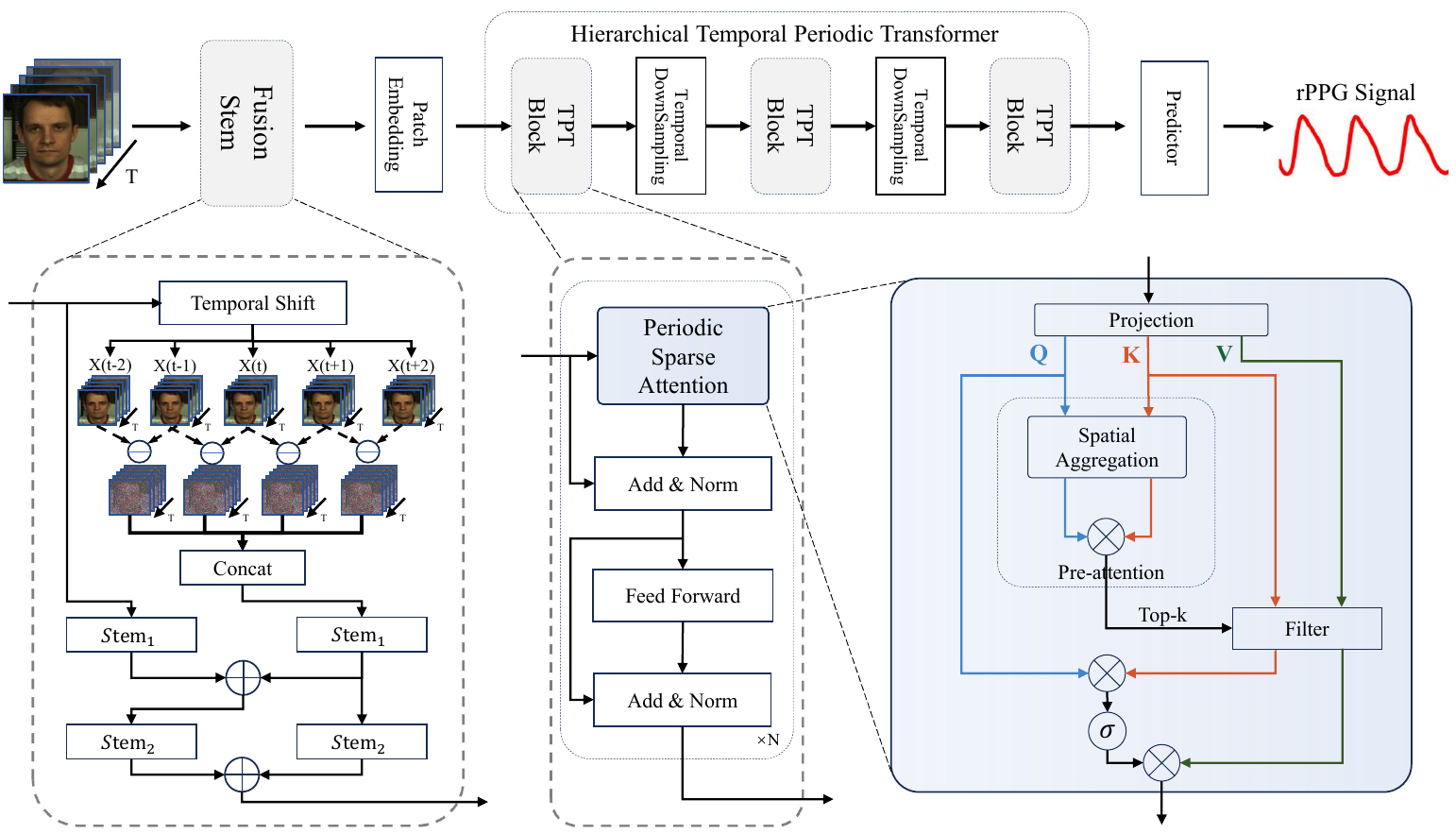}  
  \caption{The general framework of RhythmFormer. It consists of fusion stem, patch embedding, hierarchical temporal periodic transformer, and rPPG predictor head. The hierarchical temporal periodic transformer consists of three temporal periodic transformer (TPT) Blocks and two temporal downsampling modules between TPT Blocks. The attention mechanism in the TPT block is periodic sparse attention. It introduces a pre-attention stage before the vanilla attention, which selects the top-k tokens to filter out irrelevant key (K) and value (V).}
  \label{fig:2}
\end{figure*}

\section{Methodology}
In Section \ref{sec/3.1}, the general framework of RhythmFormer is presented, followed by the fusion stem in Section \ref{sec/3.2} and the hierarchical temporal periodic transformer in Section \ref{sec/3.3}. Finally, the model training is introduced in Section 3.4.

\subsection{The General Framework of RhythmFormer}
\label{sec/3.1}
As shown in Figure~\ref{fig:2}, RhythmFormer consists of fusion stem, patch embedding, hierarchical temporal periodic transformer, and rPPG prediction head (Multilayer Perceptron, MLP). We utilize the fusion stem to integrate the difference frame with the raw frame, enabling the capability to capture frame-level rPPG information. This enhances the feature representation and guides the transformer to focus on the rPPG signal. Specifically, given an RGB video input $X \in \mathbb{R}^{3 \times T \times H \times W}$, $X_{stem}=fusion\_stem(X)$, where $X_{stem} \in \mathbb{R} ^{C\times T\times H/4\times W/4}$, and C, T, W, H indicate channel, sequence length, width, and height, respectively.

Next, the output of the fusion stem will be divided into non-overlapping tokens through patch embedding. This spatially aggregates neighboring semantic information while reducing the computational cost of the transformer. Specifically, given the input  $X_{stem}$, the output of patch embedding is denoted as  $X_{token}$, where $X_{token} \in \mathbb{R} ^{C\times T\times H/16\times W/16}$. The token number is $T\times H/16\times W/16$. There is no explicit position embedding. Instead, the relative positional cues are implicitly provided through the fusion stem, the hidden layer convolution in the feed-forward, and the local context enhancement (LCE) term in the periodic sparse attention \cite{LCE}.

Subsequently, the output of patch embedding will be fed into the hierarchical temporal periodic transformer, extracting rPPG features at a fine granularity from different temporal scales through periodic sparse attention modules. It consists of three TPT Blocks with a downsampling factor of n. Depending on the different n, there is the temporal downsampling module before the TPT block. The input and output of the hierarchical temporal periodic transformer are denoted as $X_{tptin}, X_{tptout}\in R^{C\times T\times H/16\times W/16}$. Finally, the rPPG features will be projected into BVP waves through the predictor head.

\subsection{Fusion Stem}
\label{sec/3.2}
Handling the conflict between strong noise and weak features is a key issue in the rPPG task. Some methods \cite{rhythmNet,dualgan,rppgmae,zhang2025} use color space transformations based on ROI regions to obtain spatiotemporal signal maps. However, these methods ignore global contextual cues outside the predefined ROIs and require strict preprocessing steps.
There are also some methods that take raw video as input, mainly categorized into two ways. The first one is to directly use raw videos as input \cite{physnet,physformer,physformer++}. The second one is to input the normalized frame differences, while also inputting the raw frames to provide attention. The frame differences are guided by the raw frames to focus on the skin region \cite{deepphys,TSCAN,3DCDC}. The frame differences hold an advantage as it effectively removes the influence of fixed component noise (e.g., skin color and camera noise). However, other artifacts such as ambient light variation and motion artifacts still exist. This low signal-to-noise ratio poses a significant challenge for transformers based on self-attention mechanisms. When using raw frames as input, self-attention struggles to focus on the subtle changes in skin color compared with more prominent facial features. There is a higher correlation between similar spatial regions in each frame, rather than between temporally phase-similar regions. Consequently, a significant proportion of the self-attention scores originate from facially similar regions rather than capturing the rPPG features manifested by skin color changes, leading to a substantial deviation between the actual focus and the desired ones.

Based on this, we propose the fusion stem that integrates the frame differences into the raw frames, enabling frame-level representation to capture BVP wave variations. The fusion stem effectively enhances rPPG features at an extremely low computational cost, guiding the self-attention mechanism to reinforce the focus on rPPG representation. It can also significantly improve the performance of existing methods in a plug-and-play manner (see ablation study in Section \ref{sec/4.5}).

The fusion stem is inspired by the short-term TDM \cite{tdn}, but differ in motivation and implementation details. Firstly, TDM is a low-resolution architecture designed to fuse prominent motion patterns. Conversely, our fusion stem employs a fine-grained extraction approach to capture subtle inter-frame rPPG signals. Secondly, TDM utilizes residual blocks to extract motion features and integrate them into the raw frames, whereas fusion stem primarily focuses on fusing inter-frame differences without adding dedicated feature extraction modules. Instead, it employs the existing convolutions in the vanilla stem to downsample and perform preliminary feature extraction.

Specifically, as shown in Figure~\ref{fig:2}, for an input video $X \in \mathbb{R}^{3 \times T \times H \times W}$, the temporal shift is applied to obtain $X_{t-2}$, $X_{t-1}$, $X_t$, $X_{t+1}$ and $X_{t+2}$. Then, the differences between consecutive frames are computed in chronological order, resulting in $D_{t-2}$, $D_{t-1}$, $D_{t+1}$ and $D_{t+2}$. These frame differences, along with the raw frames, are passed through two separate instances of ${Stem}_1$ for primary feature extraction. ${Stem}_1$ consists of a 2D convolution layer with a kernel size of $(5\times5)$, cascaded by batch normalization (BN), ReLU, and MaxPool. The input dimension is 3 when taking the raw frames as the input and 12 when taking the concatenation of frame differences as the input.
\begin{equation}
\small
\begin{aligned} 
  &X_{diff} = Stem_1(Concat(D_{t-2},D_{t-1},D_{t+1},D_{t+2})),
  \\&X_{raw} = Stem_1(X_{t}),
  \\&X_{fusion} = X_{raw} + X_{diff}.
\end{aligned} 
\end{equation}

Then, the frame difference feature and the fused feature representation are further enhanced through $Stem_2$, which also consists of a 2D convolution layer with a kernel size of $(3\times 3)$, cascaded by BN and ReLU.
\begin{equation}
  X_{stem}= Stem_2(X_{fusion}) + Stem_2(X_{diff}).
\end{equation}

\begin{figure}[htbp]
\setlength{\abovecaptionskip}{0.0cm}
  \hspace{0.5cm} 
    \subfloat[$X_{raw}$]{\includegraphics[width=0.26\textwidth]{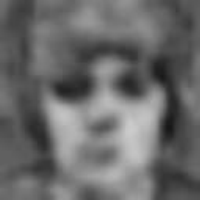}}
 \hspace{0.6cm} 
  \subfloat[$X_{diff}$]{\includegraphics[width=0.26\textwidth]{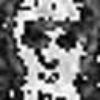}}
 \hspace{0.6cm} 
 \subfloat[Frame Difference]{\includegraphics[width=0.26\textwidth]{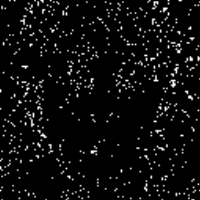}}
\caption{Visualization comparison of features extracted by the fusion stem and frame differences.}
\label{fig:fusionstem}
\vspace{-0.3cm}
\end{figure}

This design integrates multi-level features extracted from the frame differences of adjacent frames into the raw frame. Compared with the raw frame input, it significantly enhances the rPPG features, guiding the self-attention mechanism. In Figure \ref{fig:fusionstem}, the visual results of $X_{diff}$ and frame differences are compared in (b) and (c). It can be observed that the frame difference eliminates fixed component noise and focuses on subtle local color changes, but it also leads to the amplification of even the smallest noise signals, resulting in a very low signal-to-noise ratio. In contrast, the fusion stem focuses entirely on extracting frame differences from the face region. This method, which extracts deep features through convolution across adjacent five frames, contains more rPPG information and less noise compared to simple frame differences, greatly mitigating its weaknesses.

\subsection{Hierarchical Temporal Periodic Transformer}
\label{sec/3.3}
Previous studies \cite{mul1,mul3} have demonstrated the effectiveness of modeling periodic tasks with multiple temporal scales, primarily extracting temporal information through concatenation from paths at various scales. In the rPPG field, similar to many spatial multi-scale methods \cite{swintransformer,mvit}, we hypothesize that a pyramid-shaped hierarchical multi-scale learning may facilitate the provision of information from earlier learned scales to later learned scales. 

The proposed hierarchical temporal periodic transformer is composed of three TPT blocks with downsampling factors of $n$, where the values of $n$ are $1, 2, 3$ respectively. It first reinforces inter-frame rPPG features at a smaller temporal scale and then models periodic features at a larger temporal scale, effectively reducing noise interference. Specifically, for the TPT block with a downsampling factor of n, the temporal downsampling module aggregates the temporal information at a granularity of $2^n$ without overlap. The downsampling module consists of a BN layer and a 3D convolution layer with the kernel and stride size of $(2\times 1\times 1)$, resulting in an output $X_{sample}\in \mathbb{R}^{C\times {T}/{2^n}\times H/16\times W/16}$. The first TPT block does not have a dedicated downsampling module; instead, the preceding patch embedding performs the function of the downsampling module.

The first part of the TPT block is the periodic sparse attention module. Figure \ref{fig:12} illustrates the attention mechanism, where a pre-attention stage is added before the attention to ensure sufficient computational resources for fine-grained operations. The specific computation steps are shown in the blue part of Figure \ref{fig:2}.
First, the input to the module is projected. Here, Q and K are processed using the temporal difference convolution (TDC) module, while a linear projection is used to project V.
The TDC module \cite{TDC} has been proven to capture fine-grained local temporal difference features effectively \cite{autohr,3DCDC,physformer,physformer++}. Then the projected Q and K are subjected to a 3D average pooling operation to compute attention within a larger spatiotemporal receptive field, which filters out low-correlation key-value pairs. Specifically, for input $Q,K \in \mathbb{R}^{C\times{T}/{2^n}\times H/16\times W/16}$, the pooling convolutional kernel size is $(T'\times H'\times W')$.
\begin{equation}
  T'=[\frac{T}{T_s}],H'=[\frac{H/16}{4}],W'=[\frac{W/16}{4}].
\end{equation}

Where $T_s=MAX(2^x,2^n)$, and $x$ is the partition coefficient. In the experiments, since the frame rate of the dataset is uniformly downsampled to 30, $x$ is set to 2. 
The output $Q'$ and $K'$ after pooling are used to compute the attention in the pre-attention stage. The scores of the pre-attention are computed as follows:
\begin{equation}
  Score\ =\ Q'(K')^\top.
\end{equation}

The pre-attention stage is designed to obtain the indices of top-k tokens with high correlation, to filter out low-correlation key-value pairs. The pre-attention captures spatiotemporal correlations within a large receptive field, gathering the regions highly correlated with Q in both K and V. If a token from the facial area is given as a query, it corresponds to gathering facial regions with similar temporal phases to the query. The gathered key and value are expressed as:
\begin{equation}
\begin{aligned} 
      K^g\ =\ gather(K,topkIndex(Score)),
      \\V^g\ =\ gather(V,topkIndex(Score)).
\end{aligned} 
\end{equation}

The refined attention stage is computed on the top-k (high-correlation) regions identified after the pre-attention gathering. Subsequently, the local positional information is enhanced through a local context enhancement (LCE) module \cite{LCE,biformer}. LCE is a 3D convolution layer with a kernel size of $(3\times 3\times 3)$. The $i_{th}$ self-attention head output ${SA}_i$ is:
\begin{equation}
  {SA}_i\ =Softmax(\ Q{(K^g)}^\top/\sqrt d)\ V^g\ +\ LCE(V).
\end{equation}

Where $d$ represents the dimensionality of single-head attention. The output of multi-head self-attention (MHSA) is obtained by concatenating the outputs from all h attention heads, followed by a linear projection through the weight matrix $W \in \mathbb{R}^{d\times d}$.
\begin{equation}
  MHSA=Concat(SA_1,SA_2,...,SA_h)W.
\end{equation}

Like vanilla transformer \cite{transformer}, residual connection, BN, and feed-forward would be conducted after MHSA. Diverging from that, inspired by \cite{physformer,physformer++}, an additional hidden layer is incorporated into the feed-forward layer. This hidden layer consists of a 3D convolution layer, BN, and GELU, intended to refine local inconsistency and reinforce relative position cues.

\subsection{Model Training}
\label{sec/3.4}
We employed a hybrid loss function incorporating both time-domain and frequency-domain components \cite{autohr}, which can be expressed as $\mathcal{L}_{overall}=0.2\cdot\mathcal{L}_{Time}+\mathcal{L}_{Freq}$. Specifically, the temporal loss $\mathcal{L}_{Time}$ is computed using the negative Pearson correlation coefficient, and can be represented as:
\begin{equation}
\mathcal{L}_{Time}=1 -\frac{\sum_{i=1}^n \left( {S}_{gt_i} - \overline{{S}_{gt}} \right) \left( {S}_{pred_i} - \overline{{S}_{pred}} \right)}{\sqrt{\sum_{i=1}^n \left( {S}_{gt_i} - \overline{{S}_{gt}} \right)^2} \sqrt{\sum_{i=1}^n \left( {S}_{pred_i} - \overline{{S}_{pred}} \right)^2}}.
\end{equation}

Where $S$ denotes the Blood Volume Pulse waves. The frequency loss is calculated using the cross-entropy between the predicted BVP wave's frequency domain and the heart rate derived from the ground truth BVP wave. The frequency loss $\mathcal{L}_{Freq}$ is then expressed by: 
\begin{equation}
\mathcal{L}_{Freq}=CE(maxIndex(PSD({S}_{gt})),PSD({S}_{pred})).
\end{equation}  

Where $CE$ denotes cross-entropy, $maxIndex$ indicates the index of the maximum value, and $PSD$ indicates the power spectrum. 

On this basis, we further employed an HR constraint, denoted as $B_{HR}$ which serves as a bias term for selecting the optimal checkpoint. It is calculated using the distance between the ground truth HR distribution and the predicted HR distribution. The HR distribution can be represented as
$HR =N(\mu,\sigma^2)$. Where $N$ represents the normal distribution, and $\mu=maxIndex(PSD(S))$. The HR constraint is expressed by $B_{HR}\ =KL({HR}_{gt},{HR}_{pred})$. $KL$ denotes the Kullback-Leibler divergence. The final checkpoint selection is performed by $\mathcal{L}_{overall} + B_{HR}$.

%% file: sec/4_experiment.tex
\section{Experiment}
\label{sec/4_experiment}
\subsection{Dataset and Performance Metric}
Five publicly available datasets were used for evaluation: PURE~\cite{PURE}, UBFC-rPPG~\cite{UBFC-rPPG}, COHFACE~\cite{cohface}, VIPL-HR~\cite{rhythmNet}, MMPD~\cite{MMPD}. Specifically, PURE contains 60 1-minute videos comprised of 10 participants engaged in six activities, with a video resolution of 640$\times$480 and a frame rate of 30 fps. UBFC-rPPG contains 42 videos, recording subjects in a stationary state, with a resolution of 640$\times$480 and a frame rate of 30 fps. COHFACE consists of 160 one-minute video sequences from 40 participants (12 females and 28 males), with a frame rate of 20 fps. VIPL-HR includes 2,378 RGB videos from 107 participants, captured using three RGB cameras, with an unstable fps. MMPD comprised 33 participants with Fitzpatrick skin types 3-6. These participants performed four different activities (static, head rotation, speech, and walking) under four lighting conditions (LED-high, LED-low, incandescent, natural). The dataset included 660 1-minute videos with a resolution of 320$\times$240 and a frame rate of 30 fps. Notably, MMPD has two versions: a compressed mini-MMPD and an uncompressed MMPD. In this study, the mini-MMPD was chosen for experimentation. Among the five datasets, PURE and UBFC-rPPG are relatively easy, while MMPD and VIPL-HR stand out with a significantly larger sample size and more complex environment. Five commonly used metrics: Mean Absolute Error (MAE), Root Mean Squared Error (RMSE), Mean Absolute Percentage Error (MAPE), Pearson correlation coefficient ($\rho$), and Signal-to-Noise Ratio (SNR) were used for evaluation.

\subsection{Implementation Details}
We leveraged the rPPG toolbox \cite{toolbox}, an open-source toolbox based on PyTorch, to implement the proposed method. A comparative analysis was conducted against several state-of-the-art methods, all implemented with the same toolbox. 
For each video segment, the facial region was cropped and resized to 128$\times$128 pixels in the first frame, and this region was fixed in the subsequent frames. In post-processing, a second-order Butterworth filter (cutoff frequencies: 0.75Hz and 2.5 Hz) was applied to filter the predicted PPG waveform. The Welch algorithm was then employed to compute the power spectral density for further heart rate estimation. 
The value of k for top-k was set to 40. The batch size was set to 4, and the training process spanned 30 epochs. The experiments were performed using a computer that includes an AMD EPYC 7542 32-Core Processor. Network training was conducted on NVIDIA RTX A6000.

\begin{figure}[htbp]
  \centering  
  \includegraphics[width=0.99\linewidth]{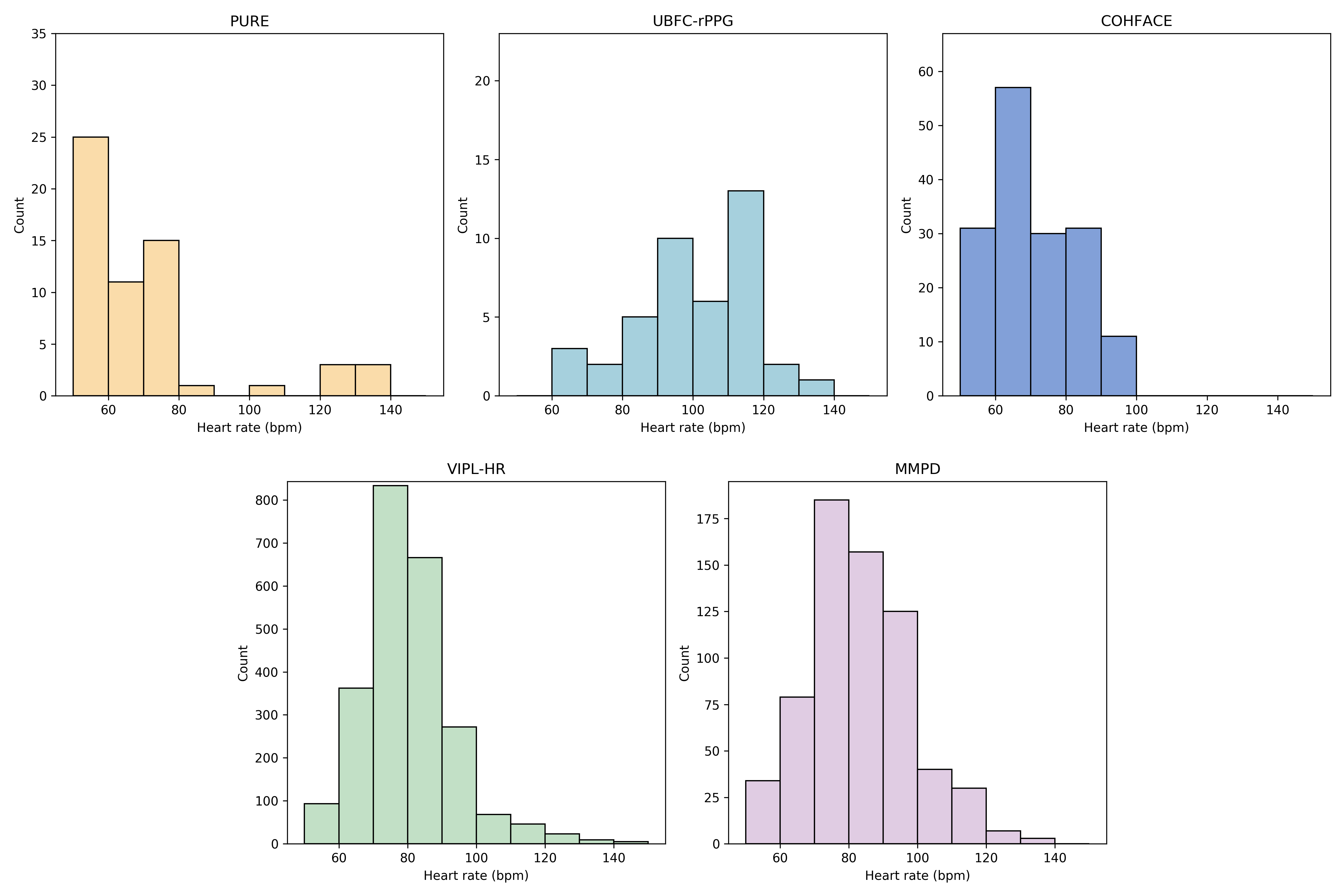}  
  \caption{Heart rate distribution among different datasets.}
  \label{fig:count}
\end{figure}

\subsection{Details of Data Augmentation}
As shown in figure \ref{fig:count}, the HR distribution among different datasets is highly imbalanced. Hence, we employed a random sampling data augmentation strategy \cite{autohr}. For cases where the heart rate, calculated from the ground truth BVP wave, exceeded 90 beats per minute (bpm), random downsampling was applied; for heart rates below 75 bpm, random upsampling was performed. Additionally, random horizontal flips were applied to all videos.

\subsection{Intra-dataset Evaluation}
\textbf{Estimation on PURE and UBFC.} For the evaluation of the PURE dataset, we adhered to the protocol outlined in \cite{HR-CNN,dualgan}. Specifically, the first 60\% of the dataset was utilized for training, while the remaining 40\% was allocated for testing. For the evaluation of the UBFC-rPPG dataset, we followed the protocol specified in \cite{pulsegan,dualgan}, selecting the initial 30 samples for training and the subsequent 12 samples for testing. According to these protocols, none of these datasets have a validation set, and the last epoch of training is used for testing. As illustrated in Table \ref{tab:intra}, on the PURE dataset, our method outperformed state-of-the-art methods across all metrics. In the case of the UBFC-rPPG dataset, our method demonstrated comparable performance to state-of-the-art methods.

\textbf{Estimation on COHFACE.} We adhered to the protocol outlined in \cite{cohface}. Specifically, the first 60\% of the dataset was utilized for training, while the remaining 40\% was allocated for testing. 
As illustrated in Table \ref{tab:intra}, our method outperforms previous approaches, achieving an MAE of 1.17. For the heavily compressed COHFACE dataset, the rPPG features may be weaker, so we conducted further exploration. Table \ref{tab:addl} presents the results using raw frames as input, processed through different stem modules. As seen in the table, the fusion stem effectively enhances rPPG features and guides attention, leading to significantly better performance with both PhysFormer and RhythmFormer.

\begin{sidewaystable}[htbp]
  \centering
  \small
  \setlength{\tabcolsep}{5.0pt}
  \renewcommand\arraystretch{0.9}
    \begin{tabular}{cccccccccccccccc}
    \toprule
          & \multicolumn{3}{c}{PURE} & \multicolumn{3}{c}{UBFC} & \multicolumn{3}{c}{COHFACE} & \multicolumn{3}{c}{VIPL-HR} & \multicolumn{3}{c}{MMPD} \\
\cmidrule{2-16}    Method & MAE$\downarrow$ & RMSE$\downarrow$ & $\rho\uparrow$ &  MAE$\downarrow$ & RMSE$\downarrow$ & $\rho\uparrow$ & MAE$\downarrow$ & RMSE$\downarrow$ & $\rho\uparrow$ &  MAE$\downarrow$ & RMSE$\downarrow$ & $\rho\uparrow$ & MAE$\downarrow$ & RMSE$\downarrow$ & $\rho\uparrow$ \\
    \midrule
    HR-CNN~\cite{HR-CNN}  & 1.84  & 2.37  & 0.98  & 4.90  & 5.89  & 0.64  & 8.10  & 10.78 & 0.29  & -     & -     & -     & -     & -     & - \\
        DeepPhys~\cite{deepphys}  & 0.83  & 1.54  & 0.99  & 6.27  & 10.82 & 0.65  & 6.89  & 13.89 & 0.34  & 11.00 & 13.80 & 0.11  & 22.27 & 28.92 & -0.03 \\
        SynRhythm~\cite{synrhythm}  & 2.71  & 4.86  & 0.98  & 5.59  & 6.82  & 0.72  &  -      &  -      &  -    & -     & -     & -     & -     & -     & - \\
    RhythmNet~\cite{rhythmNet} & -     & -     & -     & -     & -     & -     & -     & -     & -     & 5.30  & 8.14  & 0.76  & -     & -     & - \\
        PhysNet~\cite{physnet}  & 2.10  & 2.60  & 0.99  & 2.95  & 3.67  & 0.97  & 5.38  & 10.76 & -  & 10.80 & 14.80 & 0.20  & 4.80  & 11.80 & 0.60 \\
        Meta-rPPG~\cite{Meta-rppg}  & 2.52  & 4.63  & 0.98  & 5.97  & 7.42  & 0.53  & 9.31  & 12.27 & 0.19  & -     & -     & -     & -     & -     & - \\
        TS-CAN~\cite{TSCAN}  & 2.48  & 9.01  & 0.92  & 1.70  & 2.72  & 0.99  &  -      &  -      &  -    & -     & -     & -     & 9.71  & 17.22 & 0.44 \\
        CVD~\cite{CVD}  & 1.29  & 2.01  & 0.98  & 2.19  & 3.12  & 0.99  & 14.50 & 16.50 & 0.30  & 5.02  & 7.97  & 0.79  & -     & -     & - \\
        DeeprPPG~\cite{deeprppg} & 0.28  & 0.43  & 0.99  & 0.67  & 1.70  & 0.99  & 3.07  & 7.06  & 0.86  & -     & -     & -     & -     & -     & - \\
        PulseGAN~\cite{pulsegan}  & 2.28  & 4.29  & 0.99  & 1.19  & 2.10  & 0.98  &  -      &  -      &  -    & -     & -     & -     &       &       &  \\
        Dual-GAN~\cite{dualgan}  & 0.82  & 1.31  & 0.99  & 0.44  & \textbf{0.67}   & 0.99  &  -      &  -      &  -    & 4.93  & 7.68  & \textbf{0.81} & -     & -     & - \\
        AND-rPPG~\cite{AND-rPPG} &  -      &  -      &  -      & 2.67  & 4.07  & 0.92  & 6.81  & 8.06  & 0.63  & -     & -     & -     & -     & -     & - \\
        ETA-rPPGNet~\cite{etarppgnet} & 0.34  & 0.77  & 0.99  & 1.46  & 3.97  & 0.93  & 4.67  & 6.65  & 0.77  & -     & -     & -     & -     & -     & - \\
        PhysFormer~\cite{physformer}  & 1.10  & 1.75  & 0.99  & 0.50  & 0.71  & 0.99  & 16.17 & 17.92 & 0.19 & 4.97  & 7.79  & 0.78  & 11.99 & 18.41 & 0.18 \\
        EfficientPhys~\cite{efficientphys}  &  -      &  -      &  -      & 1.14  & 1.81  & 0.99  &  -      &  -      &  -    & -     & -     & -     & 13.47 & 21.32 & 0.21 \\
        TFA-PFE~\cite{AAAI23}  & 1.44  & 2.50  &  -      & 0.76  & 1.62  &  -      &  1.31 &  3.92 &  -    & -     & -     & -     & -     & -     & - \\
        SINC~\cite{sinc} & 0.61  & 1.84  & 0.99  & 0.59  & 1.83  & 0.99  &  -      &  -      &  -    & -     & -     & -     & -     & -     & - \\
        Li et al.~\cite{iccv23} & 0.64  & 1.16  & 0.99  & 0.48  & 0.64  & 0.99  &  -      &  -      &  -    & 5.19  & 8.26  & 0.78  & -     & -     & - \\
        Yue et al.~\cite{yue2023} & 1.23  & 2.01  & 0.99  & 0.58  & 0.94  & 0.99  &  -      &  -      &  -    & -     & -     & -     & -     & -     & - \\
    PhysFormer++~\cite{physformer++} & -     & -     & -     & -     & -     & -     & -     & -     & -     & 4.88  & \textbf{7.62}  & 0.80  & -     & -     & - \\
        Face2PPG$_{best}$~\cite{casado2023face2ppg} & 1.20  & 1.80  &  -      & 0.90  & 2.80  &  -      & 7.50  & 9.80  &  -    & -     & -     & -     & -     & -     & - \\
        Contrast-Phys+~\cite{contrastphys+} & 0.48  & 0.98  & 0.99  & \textbf{0.21}   & 0.80  & 0.99  &  -      &  -      &  -    & -     & -     & -     & -     & -     & - \\
        Ours   & \textbf{0.27} & \textbf{0.47} & \textbf{0.99} & 0.50  & 0.78  & 0.99  & \textbf{1.17} & \textbf{3.36} & \textbf{0.97} & \textbf{4.51} & 7.98 & 0.78  & \textbf{3.07} & \textbf{6.81} & \textbf{0.86} \\
    \bottomrule
    \end{tabular}%
  \caption{Intra-dataset results on PURE, UBFC-rPPG, COHFACE, VIPL-HR, and MMPD. The best results are in bold.}
  \label{tab:intra}%
\end{sidewaystable}%

\begin{table}[htbp]
  \centering
  \small
  \setlength{\tabcolsep}{3pt}
  \setlength{\abovecaptionskip}{0.2cm}
    \begin{tabular}{ccccccccccc}
    \toprule
          & \multicolumn{5}{c}{PhysFormer}        & \multicolumn{5}{c}{RhythmFormer} \\
\cmidrule{2-11}    Stem & MAE$\downarrow$ & RMSE$\downarrow$ & MAPE$\downarrow$ & $\rho\uparrow$ & SNR$\uparrow$  & MAE$\downarrow$ & RMSE$\downarrow$ & MAPE$\downarrow$ & $\rho\uparrow$ & SNR$\uparrow$ \\
    \midrule
    Vanilla Stem  & 16.17 & 17.92 & 25.93 & 0.19 & -11.42 & 2.05 & 5.36 & 3.33 & 0.91  & 4.76 \\
    Fusion Stem   & 2.55  & 5.81  & 4.18  & 0.90  & -5.28  & 1.17  & 3.36  & 1.84  & 0.97  & 9.28 \\
    \bottomrule
    \end{tabular}%
  \caption{Intra-dataset results trained on COHFACE.}
  \label{tab:addl}%
\end{table}%

\textbf{Estimation on VIPL-HR and MMPD}. The performance of state-of-the-art methods on the previous three datasets is nearing saturation. We further conducted an evaluation using the challenging dataset. For the VIPL-HR dataset, we followed the subject-exclusive 5-fold cross-validation protocol, as outlined in \cite{rhythmNet, physformer}. For the MMPD dataset, it is partitioned in sequence into training, validation, and test sets with a ratio of 7:1:2. As depicted in Table \ref{tab:intra}, we compared our method against existing state-of-the-art approaches. Note that the MMPD dataset is newly published. The performance of existing methods on this dataset is reimplemented by ourselves with the rPPG toolbox.
Confronted with more complex environmental conditions, our method achieves optimal performance on the MMPD dataset and shows comparable performance to state-of-the-art methods on the VIPL-HR dataset. The reason for the lower MAE but higher RMSE on VIPL-HR may be due to the presence of some extremely challenging examples in the dataset, where the head movement speed is very fast and the average frame rate is below 20. In such cases, fine-grained extraction may not provide an advantage.

\begin{figure}[htbp]
\vspace{-1.0cm}
\setlength{\abovecaptionskip}{0.0cm}
  \subfloat[]{\includegraphics[width=0.3\textwidth]{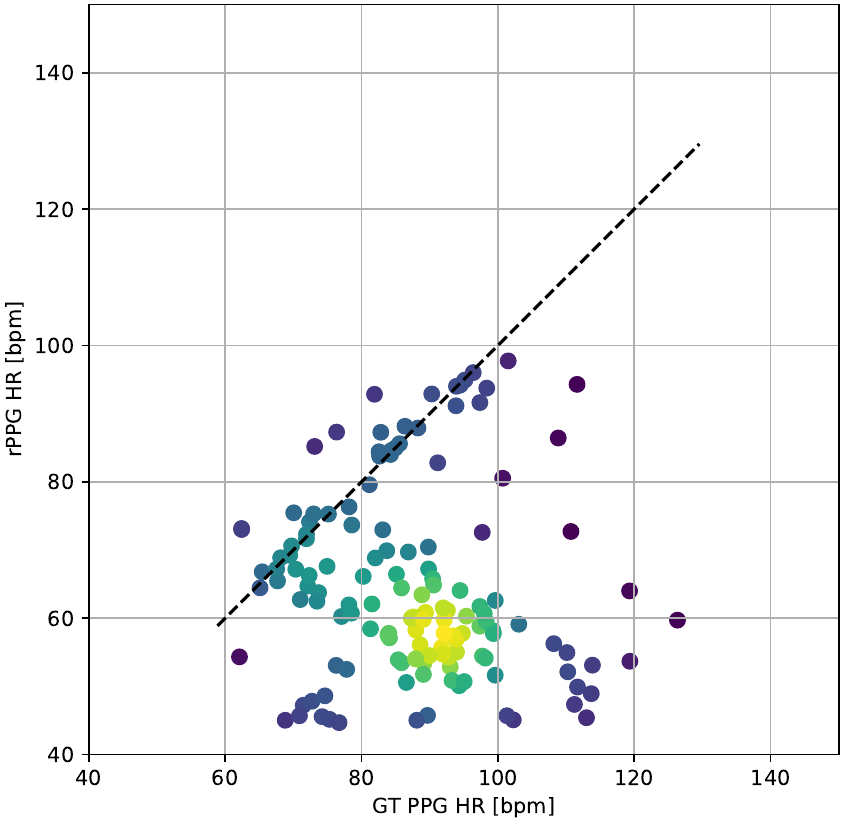}}
 \hfill 	
  \subfloat[]{\includegraphics[width=0.3\textwidth]{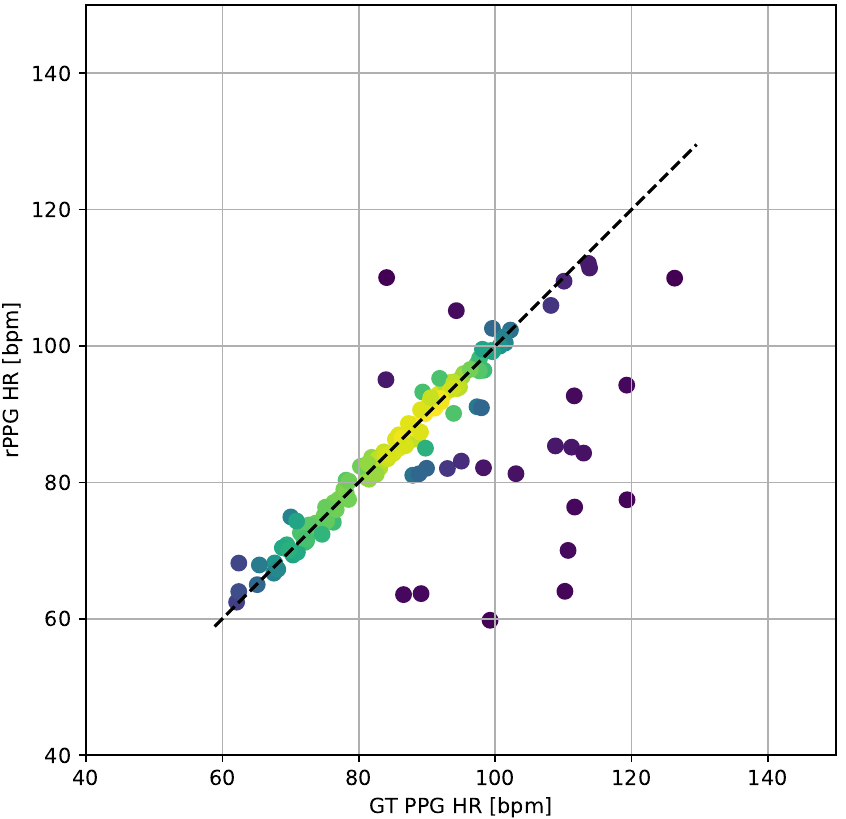}}
 \hfill	
  \subfloat[]{\includegraphics[width=0.3\textwidth]{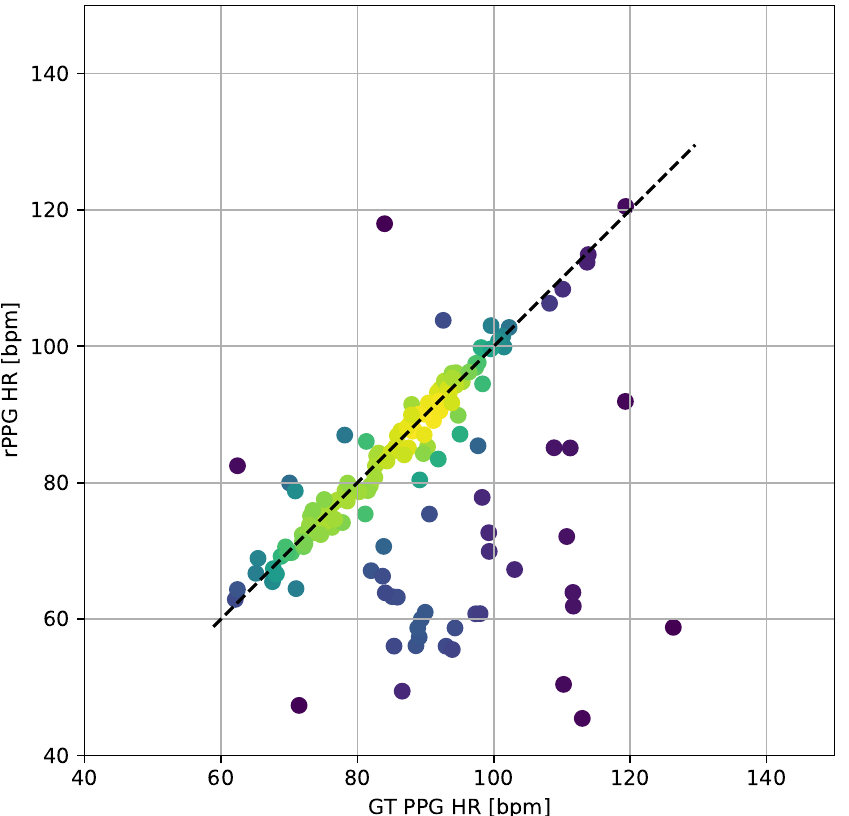}}
 \hfill 	
  \subfloat[]{\includegraphics[width=0.3\textwidth]{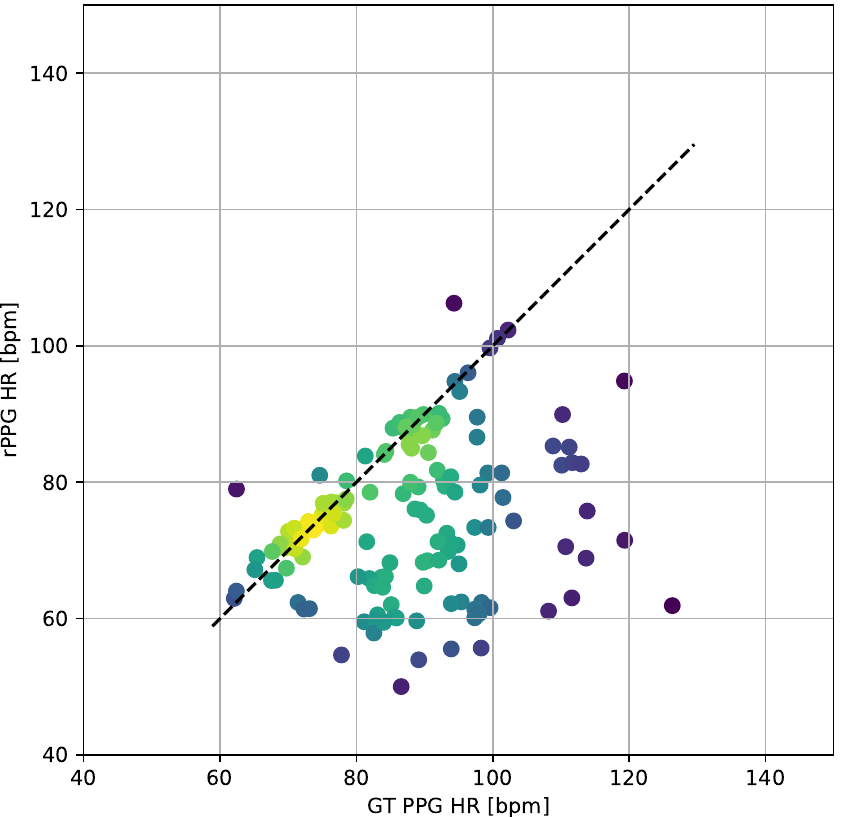}}
  \hfill	
  \subfloat[]{\includegraphics[width=0.3\textwidth]{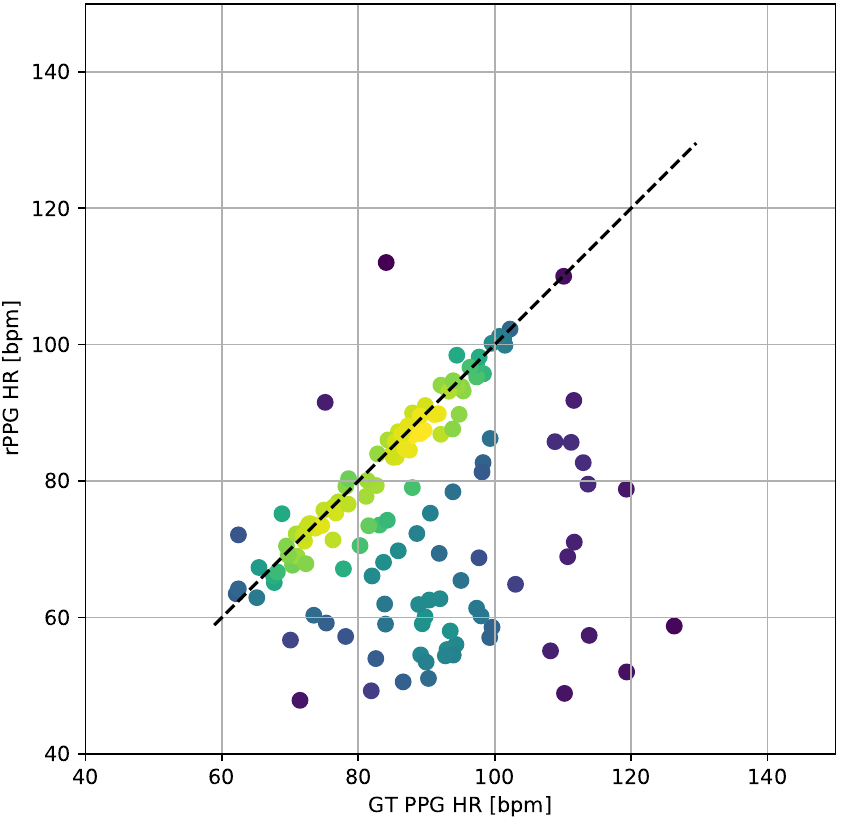}}
 \hfill 	
  \subfloat[]{\includegraphics[width=0.3\textwidth]{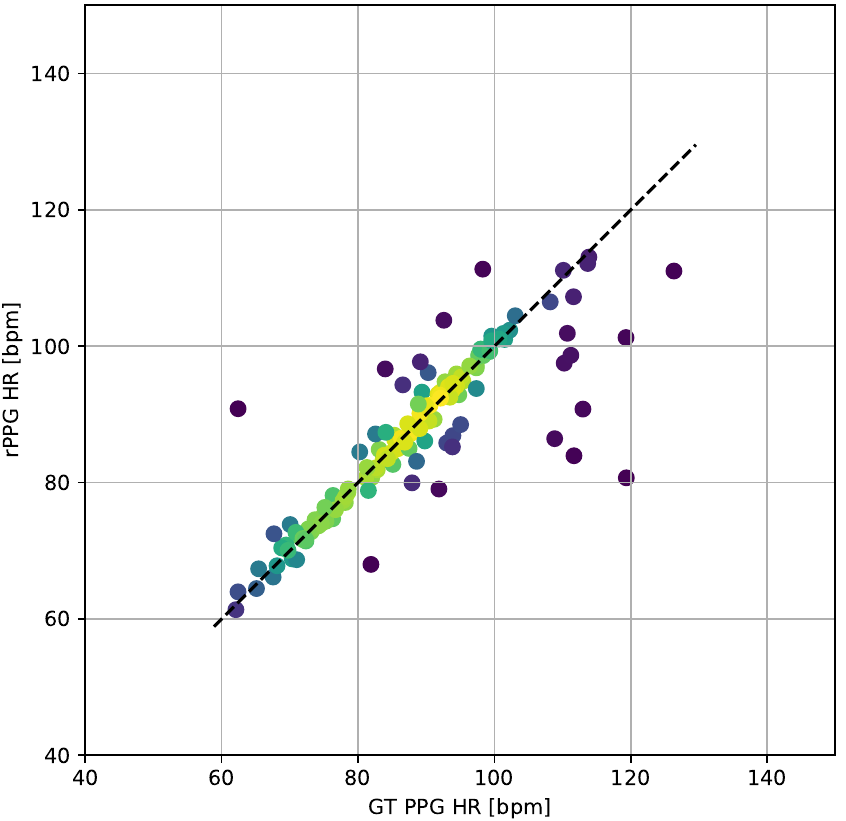}}
\caption{Scatter plots from intra-dataset evaluation on MMPD using DeepPhys~(a), PhysNet~(b), TS-CAN~(c), PhysFormer~(d), EfficientPhys~(e) and RhythmFormer~(f).}
\label{fig:mmpd1}
\end{figure}

\begin{figure}[htbp]
\vspace{-0.5cm}
\setlength{\abovecaptionskip}{0.0cm}
  \subfloat[]{\includegraphics[width=0.3\textwidth]{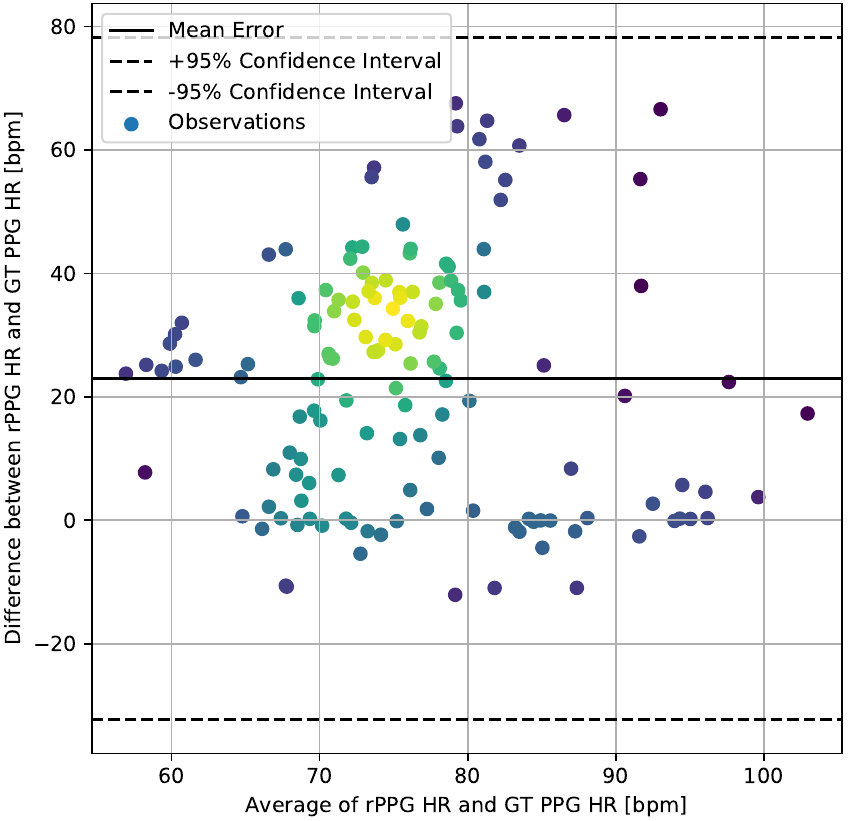}}
 \hfill 	
  \subfloat[]{\includegraphics[width=0.3\textwidth]{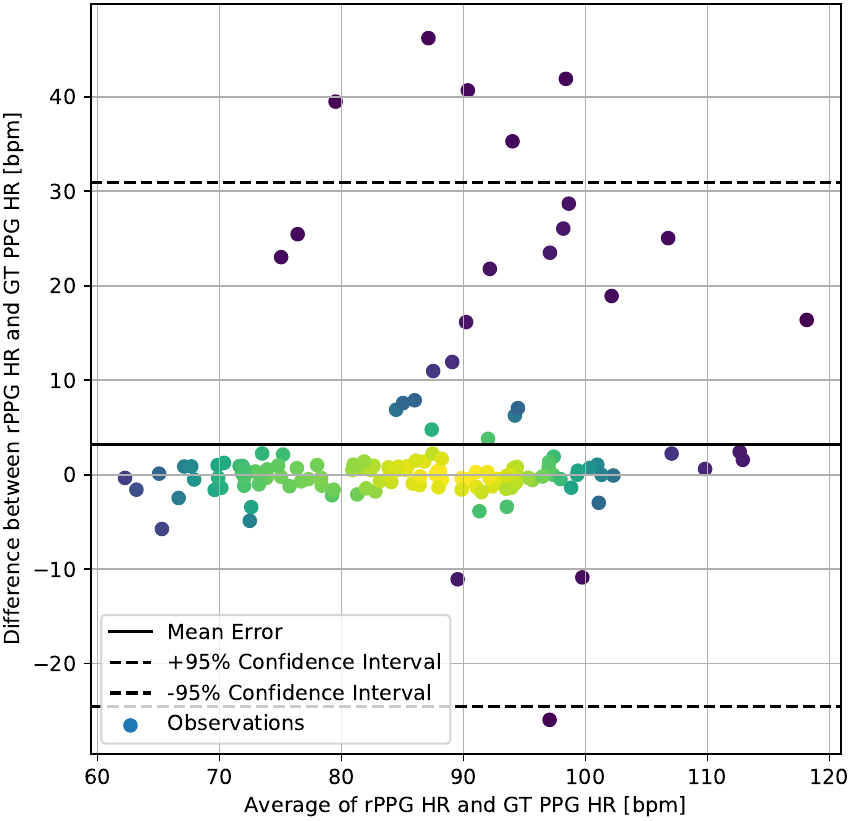}}
 \hfill	
  \subfloat[]{\includegraphics[width=0.3\textwidth]{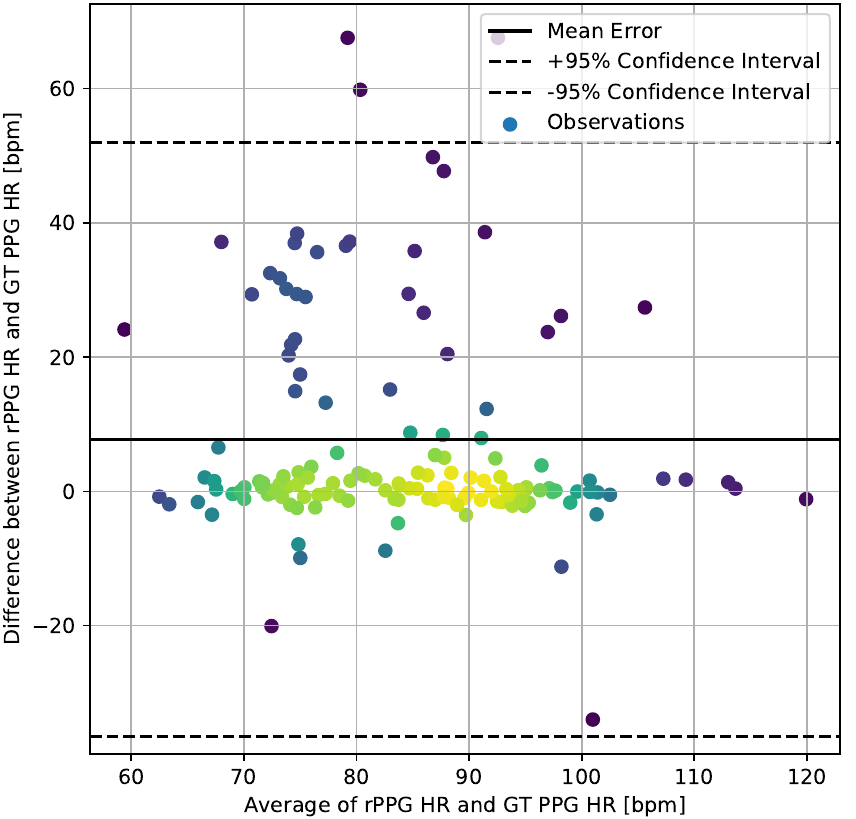}}
 \hfill 	
  \subfloat[]{\includegraphics[width=0.3\textwidth]{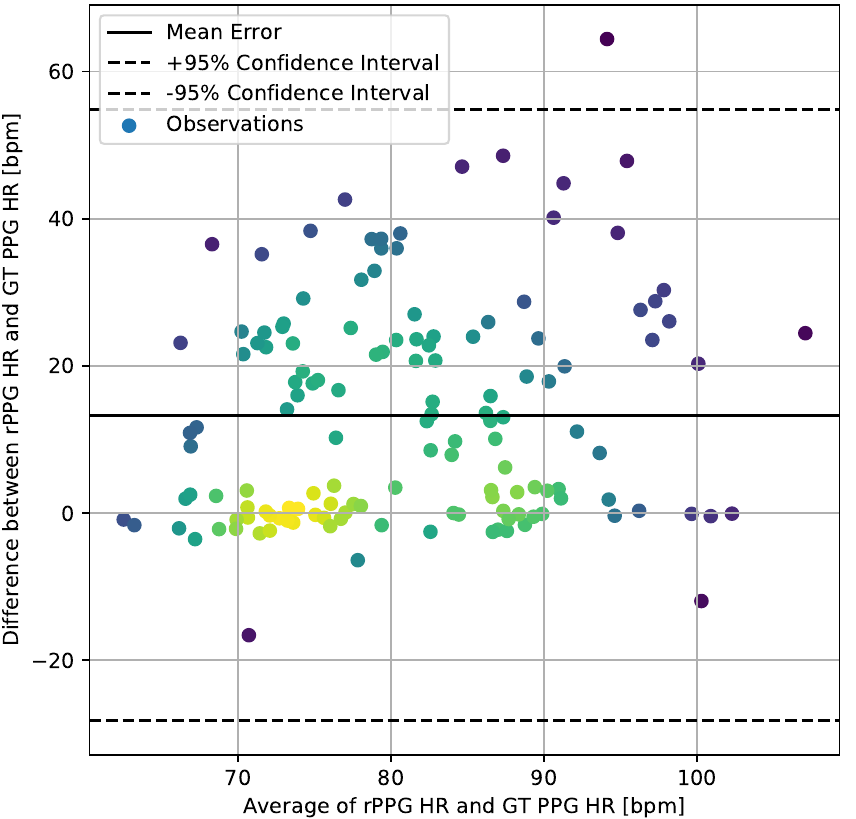}}
 \hfill	
  \subfloat[]{\includegraphics[width=0.3\textwidth]{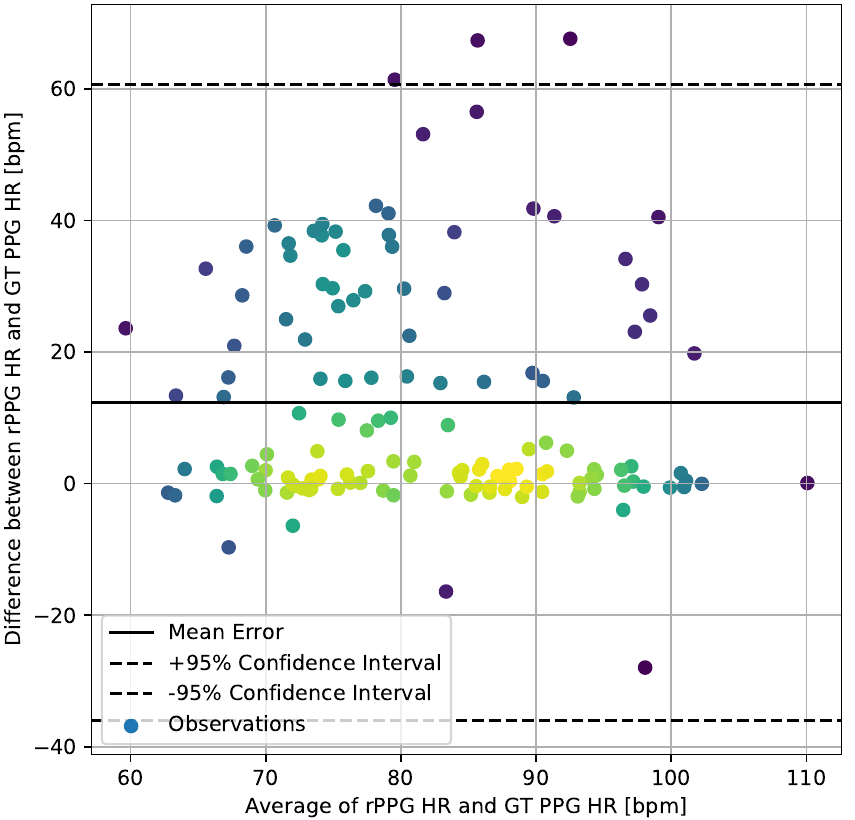}}
 \hfill 	
  \subfloat[]{\includegraphics[width=0.3\textwidth]{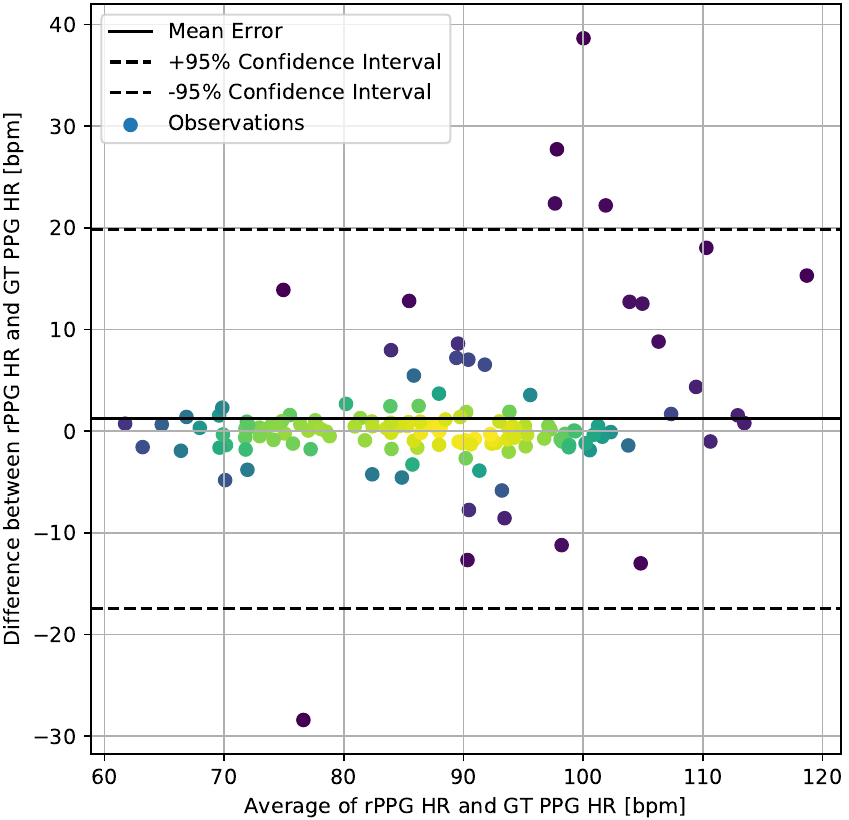}}
\caption{Bland-Altman plots from intra-dataset evaluation on MMPD using DeepPhys~(a), PhysNet~(b), TS-CAN~(c), PhysFormer~(d), EfficientPhys~(e) and RhythmFormer~(f).}
\label{fig:mmpd2}
\end{figure}

To further examine the correlation between the predicted HR and the ground truth HR, we provide scatter plots and Bland-Altman plots of intra-dataset evaluation on MMPD in Figure \ref{fig:mmpd1} and Figure \ref{fig:mmpd2}. As depicted in these figures, compared to other methods, the HR estimated by RhythmFormer shows a strong correlation with the Ground Truth HR. The scatter plot is generally aligned with the y = x line, and the points in the Bland-Altman plot mostly fall within the confidence interval, indicating the effectiveness and robustness of RhythmFormer. Meanwhile, we have noticed the presence of outliers, primarily distributed in the heart rate range of 100-110 bpm. This may be due to the distribution differences in the MMPD dataset, where the data count of heart rates exceeding 100 bpm sharply decreases (see Figure \ref{fig:count}).

\begin{figure}[htbp]
  \setlength{\abovecaptionskip}{0.0cm}
  \subfloat[]{\includegraphics[width=0.32\textwidth]{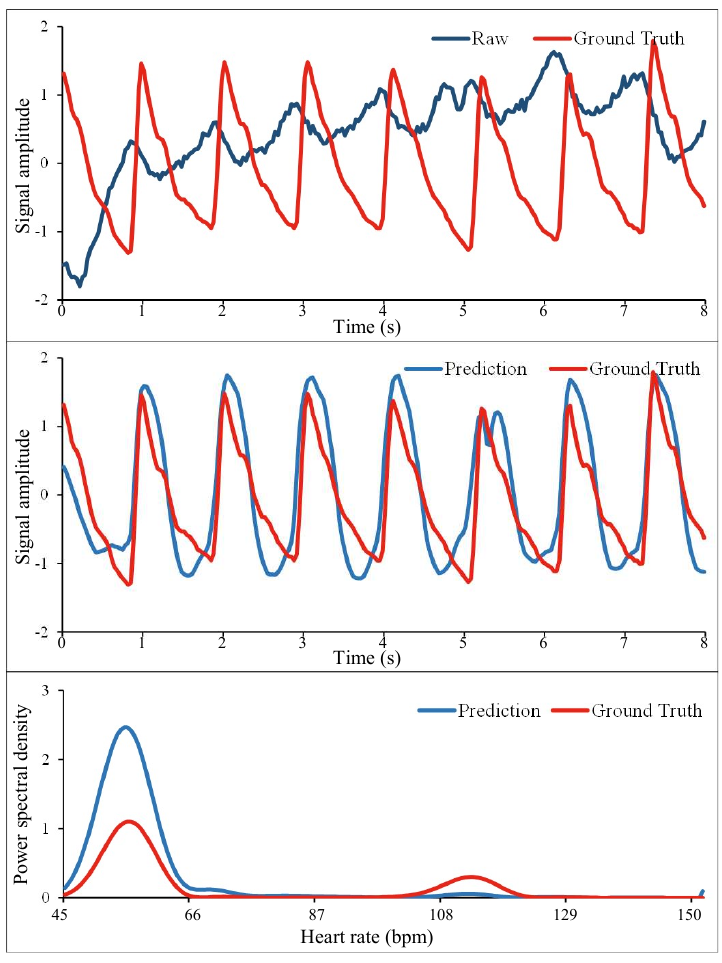}}
 \hfill	
  \subfloat[]{\includegraphics[width=0.32\textwidth]{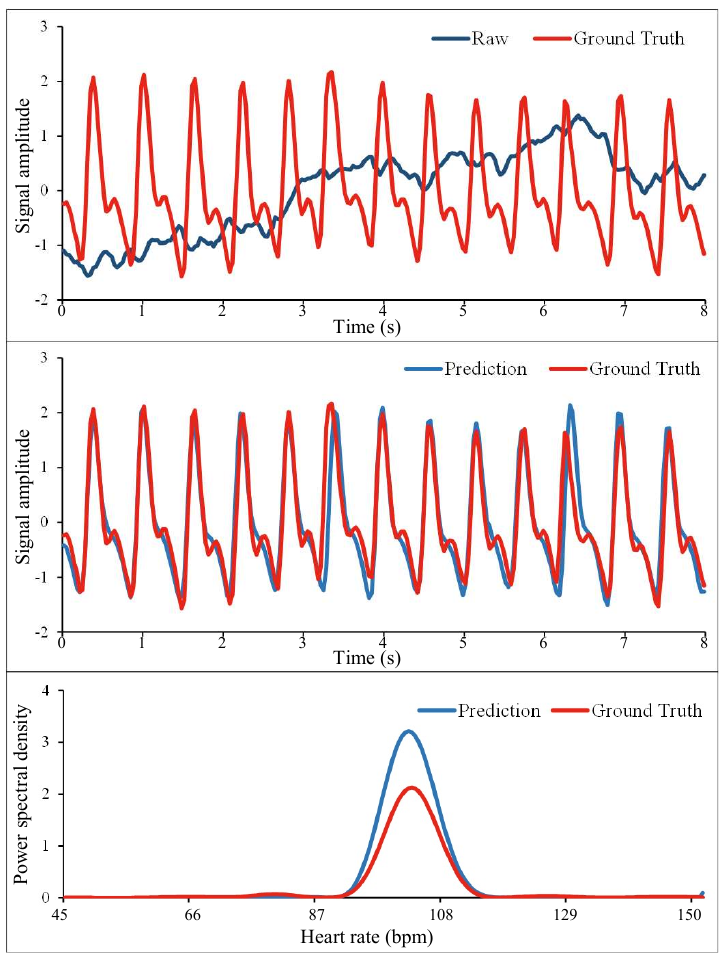}}
 \hfill	
  \subfloat[]{\includegraphics[width=0.32\textwidth]{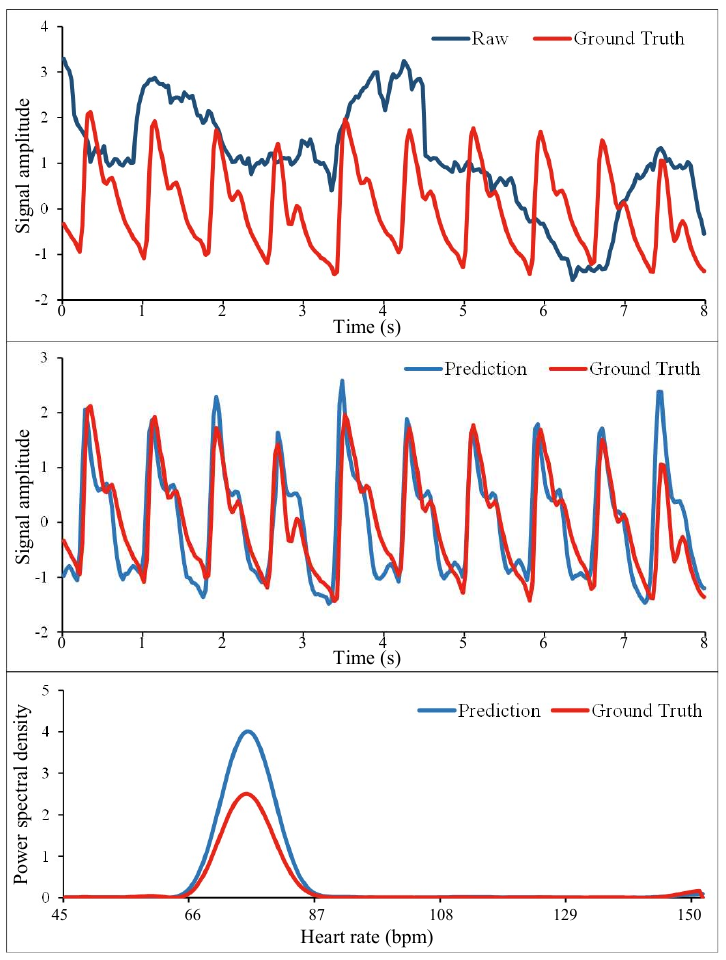}}
 \caption{Examples of HR estimation on PURE~(a), UBFC-rPPG~(b) and MMPD~(c) using RhythmFormer. “Raw” represents the raw rPPG waves (the spatial average of video frames).}
 \label{fig:example}
 \vspace{-0.1cm}
\end{figure}

Additionally, examples of the time-domain and frequency-domain plots of pulse waves predicted by RhythmFormer are provided, as illustrated in Figure \ref{fig:example}. The comparison of the raw rPPG waves, predicted rPPG signals and ground truth PPG signals in the time domain provides a direct evaluation of the model's ability to remove noise and artifacts, thereby intuitively demonstrating the model's effectiveness. The predicted signals exhibit a strong alignment with the ground truth signals in terms of trends, yet they lack detailed high-frequency information. Notably, our model robustly captures the peaks and variations of BVP signals, serving as the core judgment basis for predicting heart rate from video data.

\subsection{Cross-dataset Evaluation}
We followed the protocol outlined in \cite{toolbox} for cross-dataset evaluation. The models were trained on PURE and UBFC-rPPG individually and subsequently tested on PURE, UBFC-rPPG, and MMPD datasets. The training dataset was split, with the initial 80\% used for training and the remaining 20\% for validation. Comparative results against state-of-the-art methods are presented in Table \ref{tab:cross}. The results highlight RhythmFormer's superior capability in modeling domain-invariant rPPG features and its generalization across diverse datasets. This suggests that fine-grained feature extraction may more effectively mitigate interference from background noise and personalized facial features, enabling the model to learn domain-invariant features.

\begin{figure}[htbp]
\vspace{-0.2cm}
  \setlength{\abovecaptionskip}{0.0cm}
  \subfloat[]{\includegraphics[width=0.25\textwidth]{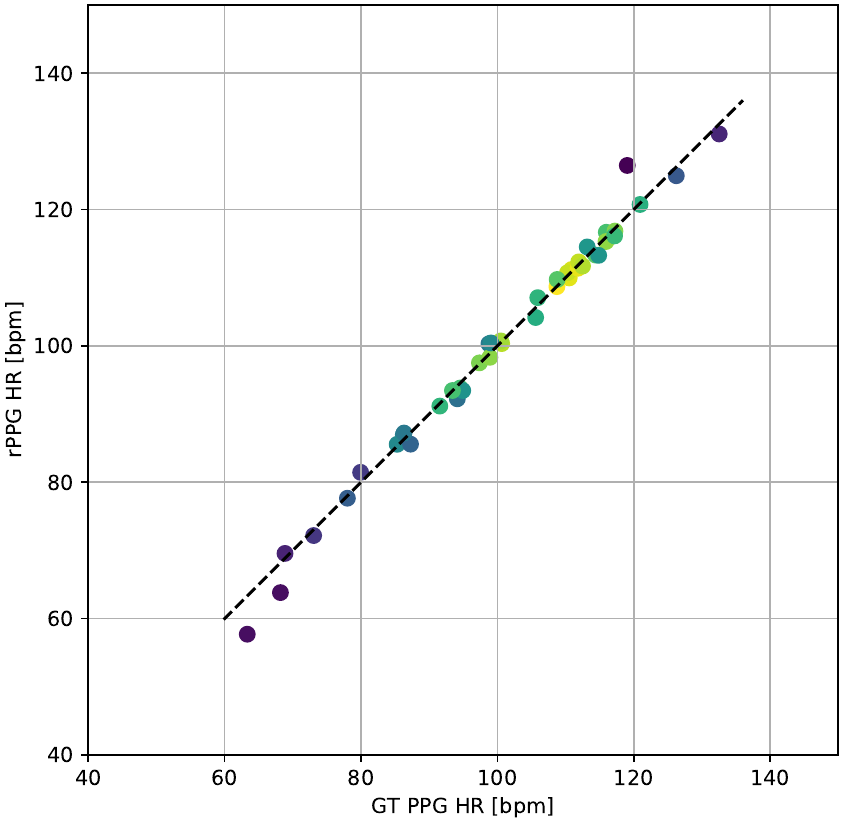}}
 \hfill 	
 \subfloat[]{\includegraphics[width=0.25\textwidth]{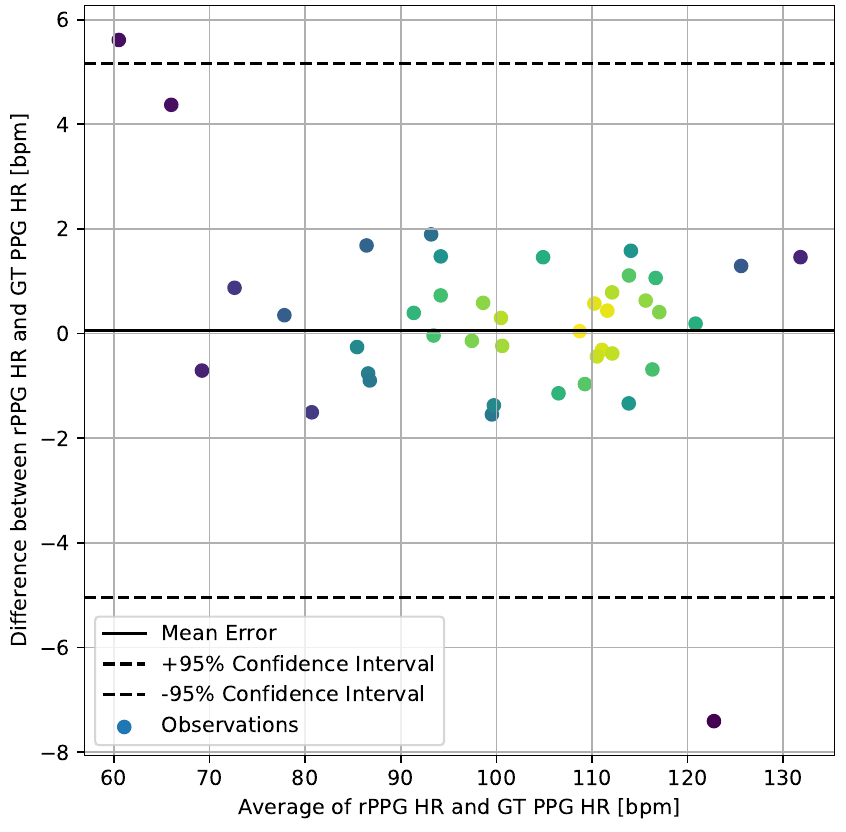}}
 \hfill 
  \subfloat[]{\includegraphics[width=0.25\textwidth]{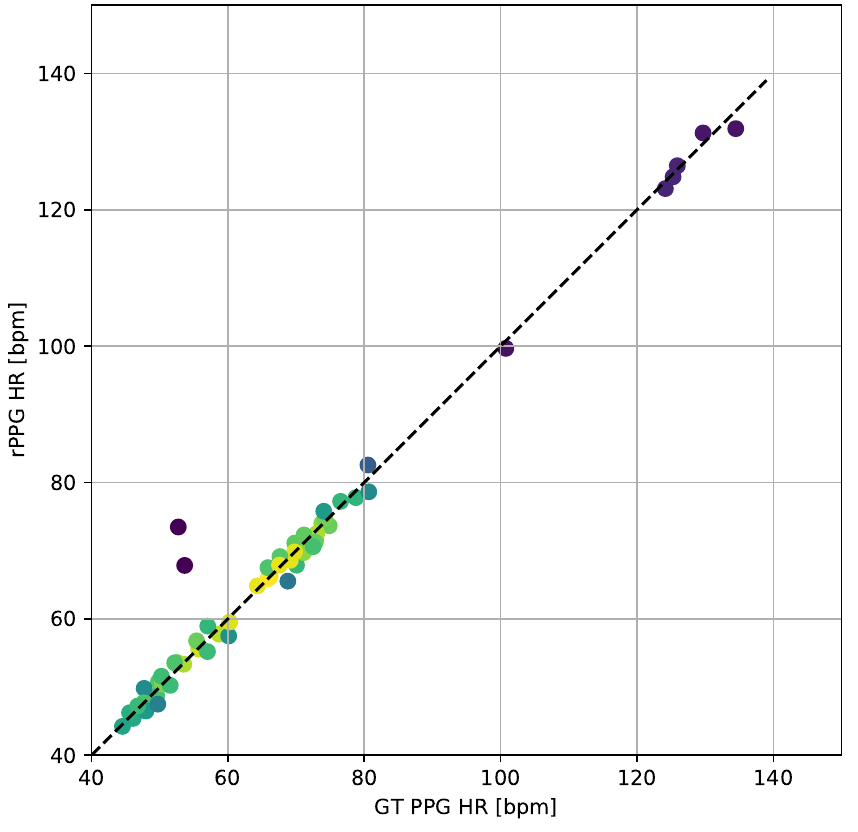}}
  \hfill 		
  \subfloat[]{\includegraphics[width=0.25\textwidth]{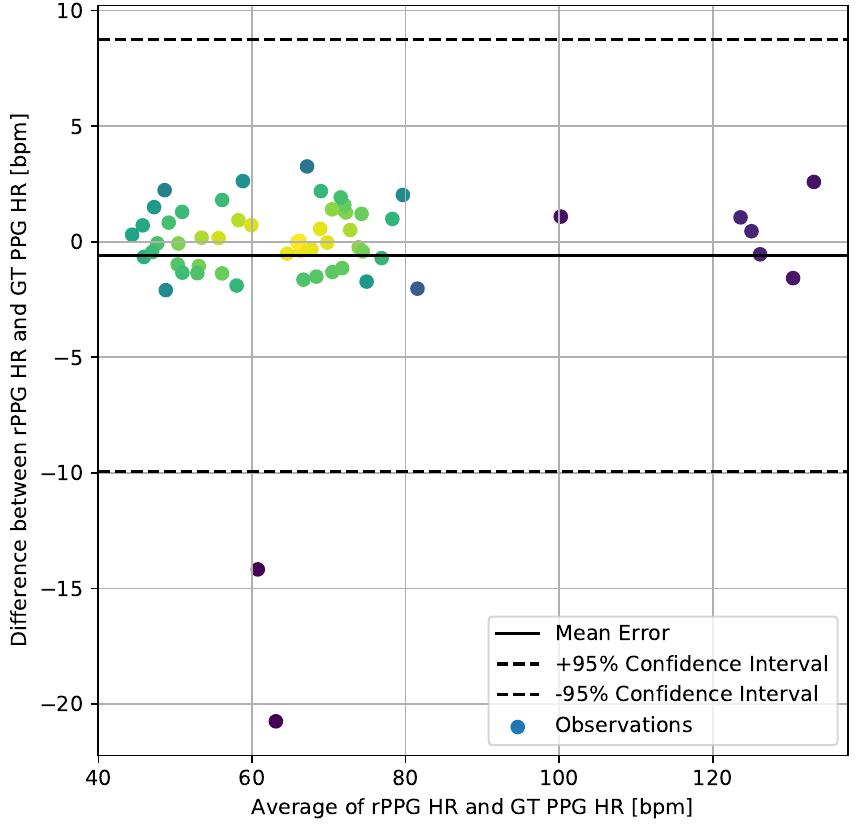}}
 \caption{Scatter plots and Bland-Altman plots of results in cross-dataset evaluation. (a, b) is training on PURE and testing on UBFC-rPPG; (c, d) is training on UBFC-rPPG and testing on PURE.}
 \label{fig:cross}
 \vspace{-0.1cm}
\end{figure}

\begin{sidewaystable}[htbp]
  \centering
  \small
  \setlength{\tabcolsep}{4pt}
  \renewcommand\arraystretch{0.92}
    \begin{tabular}{ccccccc|ccccc|ccccc}
    \toprule
          &       & \multicolumn{15}{c}{Test Set} \\
\cmidrule{3-17}          &       & \multicolumn{5}{c}{PURE}              & \multicolumn{5}{c}{UBFC-rPPG}         & \multicolumn{5}{c}{MMPD} \\
\cmidrule{3-17}    Method & Train Set & MAE$\downarrow$ & RMSE$\downarrow$ & MAPE$\downarrow$ & $\rho\uparrow$ & SNR$\uparrow$ & MAE$\downarrow$ & RMSE$\downarrow$ & MAPE$\downarrow$ & $\rho\uparrow$ & SNR$\uparrow$ & MAE$\downarrow$ & RMSE$\downarrow$ & MAPE$\downarrow$ & $\rho\uparrow$ & SNR$\uparrow$ \\
    \midrule
    GREEN~\cite{GREEN} & -     & 10.09 & 23.85 & 10.28 & 0.34  & -2.66 & 19.73 & 31.00 & 18.72 & 0.37  & -11.18 & 21.73 & 27.72 & 24.44 & -0.02 & -14.34 \\
    ICA~\cite{ICA}   & -     & 4.77  & 16.07 & 4.47  & 0.72  & 5.24  & 16.00 & 25.65 & 15.35 & 0.44  & -9.91 & 18.57 & 24.28 & 20.85 & 0.00  & -13.84 \\
    CHROM~\cite{CHROM} & -     & 5.77  & 14.93 & 11.52 & 0.81  & 4.58  & 4.06  & 8.83  & 3.84  & 0.89  & -2.96 & 13.63 & 18.75 & 15.96 & 0.08  & -11.74 \\
    LGI~\cite{LGI}   & -     & 4.61  & 15.38 & 4.96  & 0.77  & 4.50  & 15.80 & 28.55 & 14.70 & 0.36  & -8.15 & 17.02 & 23.28 & 18.92 & 0.04  & -13.15 \\
    PBV~\cite{PBV}   & -     & 3.92  & 12.99 & 4.84  & 0.84  & 2.30  & 15.90 & 26.40 & 15.17 & 0.48  & -9.16 & 17.88 & 23.53 & 20.11 & 0.09  & -13.88 \\
    POS~\cite{POS}   & -     & 3.67  & 11.82 & 7.25  & 0.88  & 6.87  & 4.08  & 7.72  & 3.93  & 0.92  & -2.39 & 12.34 & 17.70 & 14.43 & 0.17  & -11.53 \\
    Face2PPG$_{omit}$~\cite{casado2023face2ppg} & - & 4.66 & 15.82 & 4.97 & 8.76 & 4.37 & 16.99  & 29.54 & 15.91 & 0.34 & -7.29  & 17.02 & 23.23 & 18.89 & 0.04 & -12.77 \\
    \midrule
    \multirow{2}[1]{*}{DeepPhys~\cite{deepphys}} & UBFC  & 5.54  & 18.51 & 5.32  & 0.66  & 4.40  & -     & -     & -     & -     & -     & 17.50 & 25.00 & 19.27 & 0.05  & -11.72 \\
          & PURE  & -     & -     & -     & -     & -     & 1.21  & 2.90  & 1.42  & 0.99  & 1.74  & 16.92 & 24.61 & 18.54 & 0.05  & -11.53 \\
    \multirow{2}[0]{*}{PhysNet~\cite{physnet}} & UBFC  & 8.06  & 19.71 & 13.67 & 0.61  & 6.68  & -     & -     & -     & -     & -     & 10.24 & 16.54 & 12.46 & 0.29  & -8.15 \\
          & PURE  & -     & -     & -     & -     & -     & 0.98  & 2.48  & 1.12  & 0.99  & 1.09  & 13.22 & 19.61 & 14.73 & 0.23  & -10.59 \\
    \multirow{2}[0]{*}{TS-CAN~\cite{TSCAN}} & UBFC  & 3.69  & 13.80 & 3.39  & 0.82  & 5.26  & -     & -     & -     & -     & -     & 14.01 & 21.04 & 15.48 & 0.24  & -10.18 \\
          & PURE  & -     & -     & -     & -     & -     & 1.30  & 2.87  & 1.50  & 0.99  & 1.49  & 13.94 & 21.61 & 15.14 & 0.20  & -9.94 \\
    \multirow{2}[0]{*}{PhysFormer~\cite{physformer}} & UBFC  & 12.92 & 24.36 & 23.92 & 0.47  & 2.16  & -     & -     & -     & -     & -     & 12.10 & 17.79 & 15.41 & 0.17  & -10.53 \\
          & PURE  & -     & -     & -     & -     & -     & 1.44  & 3.77  & 1.66  & 0.98  & 0.18  & 14.57 & 20.71 & 16.73 & 0.15  & -12.15 \\
    \multirow{2}[0]{*}{EfficientPhys~\cite{efficientphys}} & UBFC  & 5.47  & 17.04 & 5.40  & 0.71  & 4.09  & -     & -     & -     & -     & -     & 13.78 & 22.25 & 15.15 & 0.09  & -9.13 \\
          & PURE  & -     & -     & -     & -     & -     & 2.07  & 6.32  & 2.10  & 0.94  & -0.12 & 14.03 & 21.62 & 15.32 & 0.17  & -9.95 \\
    \multirow{2}[0]{*}{Spiking-Phys.~\cite{liu2024spikingphysformer}} & UBFC  & 3.83  &  -      & 5.70  & 0.83  &  -      & -     & -     & -     & -     & -     & 14.15 &  -      & 16.22 & 0.15  &  -  \\
          & PURE  & -     & -     & -     & -     & -     & 2.80  &  -      & 2.81  & 0.95  &  -      & 14.57 &  -      & 16.55 & 0.14  &  -  \\
    \multirow{2}[1]{*}{Ours} & UBFC  &  \textbf{0.97}  &  \textbf{3.36}  &  \textbf{1.60}  &  \textbf{0.99}  &  \textbf{12.01}  & -     & -     & -     & -     & -     &  \textbf{9.08}  &  \textbf{15.07}  &  \textbf{11.17}  &  \textbf{0.53}  &  \textbf{-7.73}  \\
          & PURE  & -     & -     & -     & -     & -     &  \textbf{0.89}  &  \textbf{1.83}  &  \textbf{0.97}  &  \textbf{0.99}  &  \textbf{6.05}  &  \textbf{8.98}  &  \textbf{14.85}  &  \textbf{11.11}  &  \textbf{0.51}  &  \textbf{-8.39}  \\
    \bottomrule
    \end{tabular}%
  \caption{Cross-dataset results.}
  \label{tab:cross}%
\end{sidewaystable}%

The scatter plots and Bland-Altman plots in Figure \ref{fig:cross} further demonstrate the strong correlation between the predictions and the ground truth. It is also observed that the outliers are predominantly located in regions corresponding to the values with the least amount of data in the training set.

\subsection{Ablation Study}
\label{sec/4.5}
We conducted ablation experiments on the MMPD dataset to evaluate the impact of different modules.

\textbf{Impact of Fusion Stem.} As illustrated in Table \ref{tab:ablation1}, through the ablation study with different input forms, we validated the effectiveness of the fusion stem. Compared to raw frame input and normalized frame difference input, the fusion stem significantly improves performance. Integrating the frame differences into the raw frames enables frame-level representation to capture BVP wave variations, effectively enhancing rPPG features. This further guides the self-attention mechanism of the transformer to reinforce its focus on rPPG representation.

\begin{table}[htbp]
  \centering
  \small
  \setlength{\abovecaptionskip}{0.15cm}
    \begin{tabular}{ccccccc}
    \toprule
    Input & Stem  & MAE$\downarrow$ & RMSE$\downarrow$ & MAPE$\downarrow$ & $\rho\uparrow$ & SNR$\uparrow$  \\
    \midrule
    DiffNorm & Vanilla Stem & 3.93  & 8.66  & 3.98  & 0.79  & 3.65 \\
    Standard & Vanilla Stem & 3.56  & 7.99  & 3.71  & 0.81  & 3.52 \\
    Standard & Fusion Stem & \textbf{3.07} & \textbf{6.81} & \textbf{3.24} & \textbf{0.86} & \textbf{5.46} \\
    \bottomrule
    \end{tabular}%
  \caption{Ablation of fusion stem.}
  \label{tab:ablation1}%
\end{table}%

\begin{table}[htbp]
  \centering
  \small
  \renewcommand\arraystretch{0.9}
  \setlength{\abovecaptionskip}{0.15cm}
    \begin{tabular}{ccccccc}
    \toprule
          & \multicolumn{3}{c}{w/o Fusion Stem} & \multicolumn{3}{c}{w. Fusion Stem} \\
\cmidrule{2-7}    Backbone & MAE$\downarrow$ & RMSE$\downarrow$ & $\rho\uparrow$ & MAE$\downarrow$ & RMSE$\downarrow$ & $\rho\uparrow$ \\
    \midrule
    DeepPhys~\cite{deepphys} & 22.27 & 28.92 & -0.03 & 8.82  & 16.04 & 0.39 \\
    PhysNet~\cite{physnet} & 4.80  & 11.80 & 0.60  & 3.84  & 9.19  & 0.74 \\
    TS-CAN~\cite{TSCAN} & 9.71  & 17.22 & 0.44  & 6.91  & 13.71 & 0.51 \\
    PhysFormer~\cite{physformer} & 11.99 & 18.41 & 0.18  & 8.72  & 14.47 & 0.45 \\
    EfficientPhys~\cite{efficientphys} & 13.47 & 21.32 & 0.21  & 6.65  & 13.20 & 0.56 \\
    \bottomrule
    \end{tabular}%
  \caption{Effectiveness of fusion stem.}
  \label{tab:ablation2}%
\end{table}%

Furthermore, the fusion stem can be easily applied to other methods. We incorporated the fusion stem into existing methods. As demonstrated in Table \ref{tab:ablation2} and Table \ref{tab:addl}, the application of the fusion stem to other methods achieved consistent performance improvements. This strongly attests to the effectiveness of the fusion stem, showcasing its capability to enhance rPPG features and improve signal-to-noise ratio across different methods.

We conducted a detailed ablation study on the fusion stem. Table \ref{tab:fn} reports the ablation results based on the number of adjacent frames, while another part of Table \ref{tab:fs} reports the ablation results based on the step of adjacent frames. The best performance was achieved when using two surrounding frames with a step of 1. Steps of 0.5 and 1.5 involved cubic spline interpolation, which may explain the slight performance drop. The ablation results from the frame step also indicate that small variations in fps do not significantly affect the model's performance, which is consistent with the results from the COHFACE dataset in Table \ref{tab:fps}.

\begin{table}[htbp]
  \centering
  \small
  \setlength{\tabcolsep}{10pt}
  \renewcommand\arraystretch{0.95}
  \setlength{\abovecaptionskip}{0.1cm}
    \begin{tabular}{cccccc}
    \toprule
    Frame Number & MAE$\downarrow$ & RMSE$\downarrow$ & MAPE$\downarrow$ & $\rho\uparrow$ & SNR$\uparrow$  \\
    \midrule
    0     & 4.56  & 9.44  & 4.60  & 0.73  & 0.68 \\
    1     & 4.19  & 8.90  & 4.29  & 0.77  & 1.94 \\
    2     & \textbf{3.07} & \textbf{6.81} & \textbf{3.24} & \textbf{0.86} & \textbf{5.46} \\
    3     & 3.74  & 8.24  & 3.86  & 0.79  & 4.26 \\
    \bottomrule
    \end{tabular}%
  \caption{Ablation of frame number.}
  \label{tab:fn}%
\end{table}%

\begin{table}[htbp]
  \centering
  \small
  \setlength{\tabcolsep}{10pt}
  \renewcommand\arraystretch{0.95}
  \setlength{\abovecaptionskip}{0.1cm}
    \begin{tabular}{cccccc}
    \toprule
    Frame Step & MAE$\downarrow$ & RMSE$\downarrow$ & MAPE$\downarrow$ & $\rho\uparrow$ & SNR$\uparrow$  \\
    \midrule
    0.5   & 4.24  & 9.47  & 4.40  & 0.73  & 3.47 \\
    1.0   & \textbf{3.07} & \textbf{6.81} & \textbf{3.24} & \textbf{0.86} & \textbf{5.46} \\
    1.5   & 3.45  & 8.19  & 3.47  & 0.80  & 4.45 \\
    2.0   & 3.15  & 6.51  & 4.28  & 0.79  & 4.60 \\
    \bottomrule
    \end{tabular}%
  \caption{Ablation of frame step.}
  \label{tab:fs}%
\end{table}%

\begin{table}[htbp]
  \centering
  \small
  \setlength{\tabcolsep}{12pt}
  \renewcommand\arraystretch{0.8}
  \setlength{\abovecaptionskip}{0.1cm}
    \begin{tabular}{cccccc}
    \toprule
    Fusion Ratio & MAE$\downarrow$ & RMSE$\downarrow$ & MAPE$\downarrow$ & $\rho\uparrow$ & SNR$\uparrow$ \\
    \midrule
    $0 : 10$ & 4.39  & 9.10  & 4.55  & 0.76  & 4.08 \\
    $1 : 9$ & 3.45  & 7.77  & 3.56  & 0.82  & 5.16 \\
    $3 : 7$ & 3.85  & 8.13  & 4.02  & 0.80  & 4.41 \\
    $5 : 5$ & \textbf{3.07} & \textbf{6.81} & \textbf{3.24} & \textbf{0.86} & \textbf{5.46} \\
    $7 : 3$ & 3.36  & 7.99  & 3.46  & 0.81  & 4.42 \\
    $9 : 1$ & 3.27  & 7.12  & 3.43  & 0.85  & 4.71 \\
    $10 : 0$ & 4.56  & 9.44  & 4.60  & 0.73  & 0.68 \\
    \bottomrule
    \end{tabular}%
  \caption{Ablation of fusion ratio}
  \label{tab:ratio}%
\end{table}%

Table \ref{tab:ratio} presents the ablation study of the fusion ratio when integrating the raw frames with the frame differences. The optimal results are achieved with a 5:5 fusion ratio. Furthermore, the first and last rows indicate that the fusion of raw frames and frame differences significantly enhances the rPPG feature representation, thereby improving the recovery quality of the BVP waveform.

\textbf{Impact of Periodic Sparse Attention.} As illustrated in Table \ref{tab:ablation3}, the result demonstrates that, compared to the vanilla attention mechanism, the periodic sparse attention mechanism achieves a finer-grained context relevance computation through pre-attention, significantly enhancing the modeling capability while reducing noise interference.

\begin{table}[htbp]
  \centering
  \small
  \setlength{\abovecaptionskip}{0.15cm}
    \begin{tabular}{cccccc}
    \toprule
    Module & MAE$\downarrow$ & RMSE$\downarrow$ & MAPE$\downarrow$ & $\rho\uparrow$ & SNR$\uparrow$ \\
    \midrule
    Vanilla Attention  & 4.43  & 9.08  & 4.53  & 0.76  & 3.30 \\
    Periodic Sparse Attention  & \textbf{3.07} & \textbf{6.81} & \textbf{3.24} & \textbf{0.86} & \textbf{5.46} \\
    \bottomrule
    \end{tabular}%
  \caption{Ablation of periodic sparse attention.}
  \label{tab:ablation3}%
\end{table}%

\begin{table}[htbp]
  \centering
  \small
  \renewcommand\arraystretch{0.75}
  \setlength{\abovecaptionskip}{0.1cm}
    \begin{tabular}{cccccccc}
    \toprule
    TPT Number & Strategy & n  & MAE$\downarrow$ & RMSE$\downarrow$ & MAPE$\downarrow$ & $\rho\uparrow$ & SNR$\uparrow$  \\
    \midrule
    1     & -     & 0     & 4.52  & 10.24 & 4.88  & 0.69  & 0.89 \\
    1     & -     & 1     & 4.71  & 10.63 & 4.97  & 0.67  & 0.93 \\
    1     & -     & 2     & 3.75  & 8.01  & 3.86  & 0.81  & 2.83 \\
    1     & -     & 3     & 4.26  & 8.85  & 4.42  & 0.76  & 3.57 \\
    2  & hierarchical & 1,2   & 4.07  & 8.86  & 4.19  & 0.77  & 4.07 \\
    3     & concat & 1,2,3 & 3.61  & 7.25  & 3.83  & 0.85  & 0.72 \\
    3     & hierarchical & 3,2,1 & 5.10   & 10.73 & 5.21  & 0.64  & 0.96 \\
    3     & hierarchical & 1,2,3 & \textbf{3.07} & \textbf{6.81} & \textbf{3.24} & \textbf{0.86} & \textbf{5.46} \\
    \bottomrule
    \end{tabular}%
  \caption{Ablation of multi-temporal scale.}
  \label{tab:ablation4}%
\end{table}%

\textbf{Impact of Multi-temporal Scale.} We compared the performance of the model under different strategies and at different temporal scales, as shown in Table \ref{tab:ablation4}. Here, n represents the downsampling factor of the TPT block. The results in the first four rows indicate that, compared to a single temporal scale, multi-scale modeling can more effectively capture the periodic rPPG features.

 Since the frame rate is 30 fps, when the number of TPT blocks exceeds 3, the downsampling factor becomes too large to capture a single heartbeat at relatively high heart rates. Therefore, the ablation study was conducted only for TPT block values up to 3.
The results from the $2_{nd}$, $5_{th}$, and $8_{th}$ rows demonstrate a continuous improvement in performance as the scale increases. This may be due to the varying effects of noise at different scales, and the multi-scale approach can reduce the weight of noise at a single scale. We have plotted the Bland-Altman plots for these three rows in Figure 10 and also provided the specific waveforms for the same outlier sample. It can be observed that the distributions of the three Bland-Altman plots are quite similar, but the multi-scale approach effectively mitigates some outliers by reducing noise, leading to a smaller 95\% confidence interval. From the waveforms, it is clear that the multi-scale method enhances important features while suppressing noise.

\begin{figure}[htbp]
\setlength{\abovecaptionskip}{0.1cm}
 \subfloat[Table 10, $2_{nd}$ row]{\includegraphics[width=0.33\textwidth]{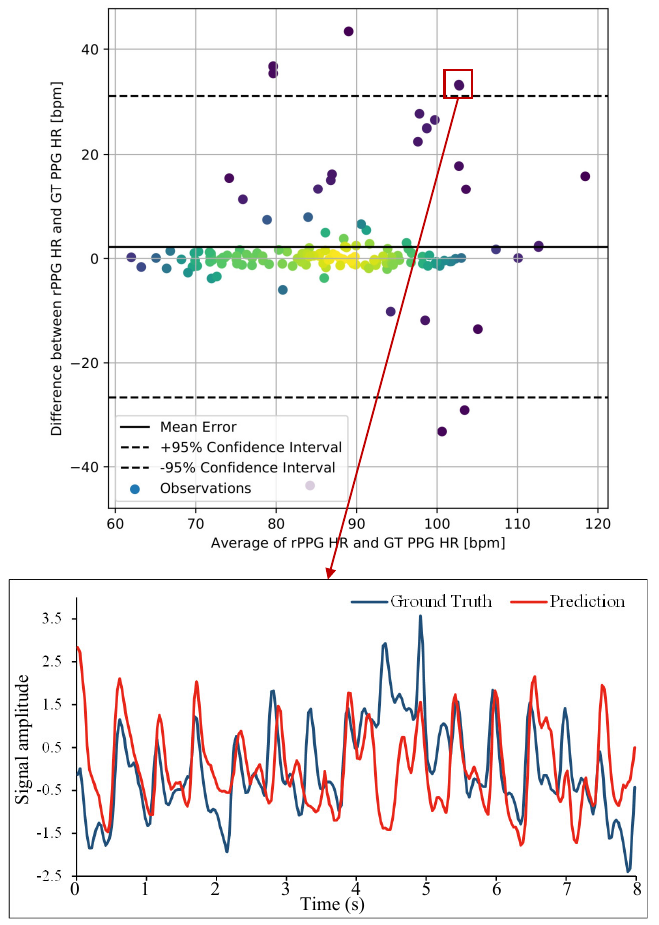}}
 \hfill	
  \subfloat[Table 10, $5_{th}$ row]{\includegraphics[width=0.33\textwidth]{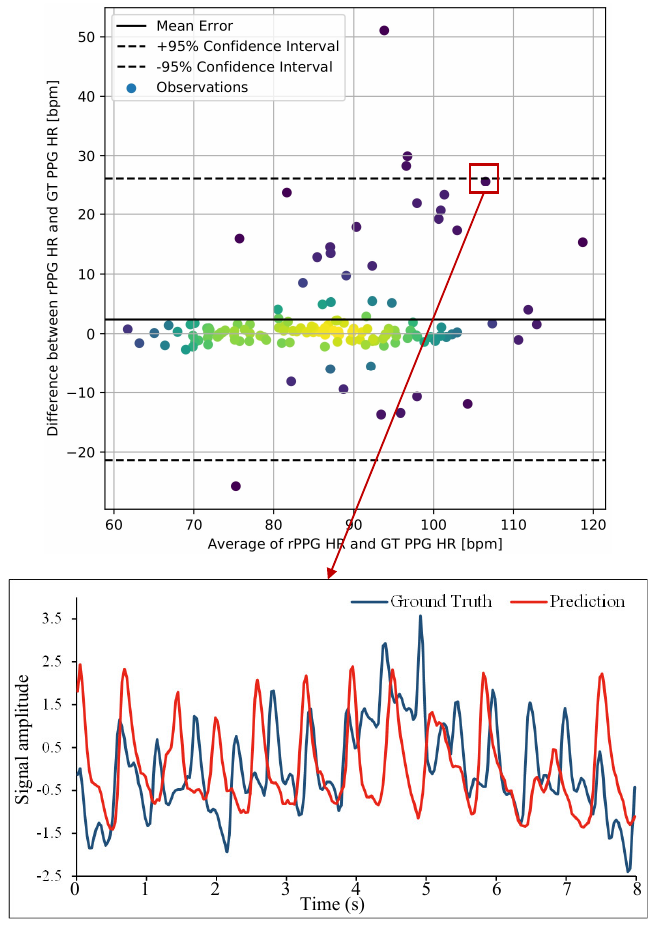}}
 \hfill	
  \subfloat[Table 10, $8_{th}$ row]{\includegraphics[width=0.33\textwidth]{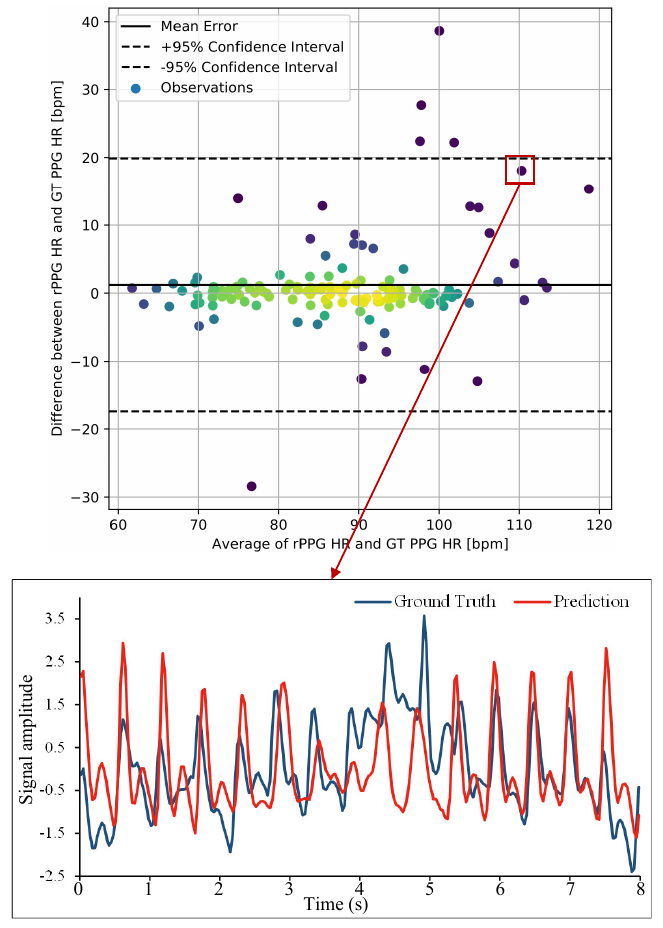}}
\caption{Visualization of the ablation study on multi-temporal scale.}
\label{fig:hiera}
\end{figure}

The results in the $6_{th}$ row suggest that, unlike other periodic tasks (such as action counting \cite{mul1}), hierarchical feature extraction is more effective than concatenation for extracting rPPG signals, which belong to a single category and exhibit quasi-periodicity. Meanwhile, the comparison between the last two rows shows that the multi-scale approach with increasing temporal granularity significantly outperforms the approach with decreasing temporal granularity. 
This may be because noise typically manifests as high-frequency signals, and fine-grained features are more susceptible to noise interference. The abstract process from fine to coarse naturally ignores detailed noise and focuses on semantic information. The pyramid structure uses higher-level features (global context) to guide the optimization of lower-level features, enhancing important features and suppressing noise. As shown in the $5_{th}$ row, although the heart rate derived from the pulse wave is relatively inaccurate, the SNR of the reconstructed BVP compared to the ground truth BVP is relatively high, which supports this point.

It is noteworthy that, for a single temporal scale, a granularity of 4-frame (n=2) is significantly superior to others. Particularly, when the scale is set to 1-frame (n=0), i.e., globally attending to the temporal sequence, it incurs substantial memory and computational costs, yet the effect is suboptimal. This suggests that aggregating temporal information effectively enhances rPPG features and prevents noise interference.

\textbf{Impact of Input Length.} As shown in Table \ref{tab:length}, an input length of 160 frames achieves the optimal results. This may be because face detection is performed only on the first frame. If the input length is too long, subsequent frames may lose the face due to motion in certain scenarios. In contrast, shorter input lengths may not effectively capture inter-frame rPPG signals.

\begin{table}[ht]
  \centering
  \small
    \setlength{\tabcolsep}{10pt}
  \renewcommand\arraystretch{0.8}
  \setlength{\abovecaptionskip}{0.05cm}
    \begin{tabular}{cccccc}
    \toprule
    Input Length & MAE$\downarrow$ & RMSE$\downarrow$ & MAPE$\downarrow$ & $\rho\uparrow$ & SNR$\uparrow$  \\
    \midrule
    80    & 3.53  & 7.77  & 3.64  & 0.82  & 2.74 \\
    160   & \textbf{3.07} & \textbf{6.81} & \textbf{3.24} & \textbf{0.86} & 5.46 \\
    240   & 3.34  & 7.12  & 3.57  & 0.85  & \textbf{6.03} \\
    320   & 3.86  & 7.81  & 4.09  & 0.83  & 4.15 \\
    \bottomrule
    \end{tabular}%
  \caption{Ablation of input length.}
  \label{tab:length}%
\end{table}%

\textbf{Impact of FPS on the COHFACE.} 
The original frame rate of the COHFACE dataset is 20 fps. Here, we conducted a comparative experiment by upsampling to 30 fps. As shown in Table \ref{tab:fps}, after upsampling, the MAE, RMSE, and SNR deteriorated, while the MAPE and $\rho$ improved. This indicates that upsampling allows the model to better capture the overall variation patterns but also introduces noise, reducing the model's robustness.

\begin{table}[ht]
  \centering
  \setlength{\tabcolsep}{12pt}
  \renewcommand\arraystretch{0.8}
  \setlength{\abovecaptionskip}{0.05cm}
  \small
    \begin{tabular}{cccccc}
    \toprule
    FPS & MAE$\downarrow$ & RMSE$\downarrow$ & MAPE$\downarrow$ & $\rho\uparrow$ & SNR$\uparrow$  \\
    \midrule
    20 & 1.17  & 3.36  & 1.84  & 0.97  & 9.28 \\
    30 & 1.54  & 3.60  & 1.51  & 0.98  & 5.61 \\
    \bottomrule
    \end{tabular}%
  \caption{Impact of FPS on the COHFACE.}
  \label{tab:fps}%
\end{table}%

\textbf{Impact of Loss Term and Bias Term.} 
We conducted ablation experiments on each loss term and bias term, and the results indicate that the temporal loss is the core component, while the frequency-domain loss and HR bias also play a role.

\begin{table}[ht]
  \centering
  \small
  \renewcommand\arraystretch{0.8}
  \setlength{\abovecaptionskip}{0.05cm}
    \begin{tabular}{cccccccc}
    \toprule
    $\mathcal{L}_{Time}$ & $\mathcal{L}_{Freq}$ & $B_{HR}$ & MAE$\downarrow$ & RMSE$\downarrow$ & MAPE$\downarrow$ & $\rho\uparrow$ & SNR$\uparrow$  \\
    \midrule
    \Checkmark     &     &      & 3.56  & 8.17  & 3.61  & 0.80  & 4.19 \\
    &     \Checkmark     &     & 13.32 & 16.54 & 14.15 & 0.38  & -12.7 \\
    \Checkmark     & \Checkmark     &      & 3.13  & 6.98  & 3.39  & 0.85  & \textbf{5.56} \\
    \Checkmark & \Checkmark & \Checkmark & \textbf{3.07} & \textbf{6.81} & \textbf{3.24} & \textbf{0.86} & 5.46 \\
    \bottomrule
    \end{tabular}%
  \caption{Ablation of loss term and bias term.}
  \label{tab:loss}%
\end{table}%

\subsection{Computational Cost}
We conducted a 30-second inference at a resolution of 128$\times$128 for all implemented methods, reporting the parameter count, average multiply-accumulates (MACs) per frame, average throughput per frame, and average peak GPU memory per frame. The notation ``(10s)" refers to the average results for inferring 10 seconds. We found that when inferring 10 seconds, the average computational cost per frame of CNN-based methods remained unchanged, while the throughput and GPU memory usage of transformer-based methods varied. GPU memory usage significantly increased due to the pairwise attention computation for tokens in transformer-based methods. The throughput analysis is provided in Section 4.9.

\begin{table}[htbp]
  \centering
  \renewcommand\arraystretch{0.8}
  \setlength{\tabcolsep}{3pt}
  \setlength{\abovecaptionskip}{0.15cm}
  \small
    \begin{tabular}{ccccc}
    \toprule
    Methods & Parameters (M) & MACs (M) & Throughput(Kfps) & GPU Memory(M) \\
    \midrule
    DeepPhys~\cite{deepphys} & 1.98  & 744.45 & 5.65  & 37.28 \\
    PhysNet~\cite{physnet} & 0.77  & 438.24 & 11.80 & 11.43 \\
    TS-CAN~\cite{TSCAN} & 1.98  & 744.45 & 5.21  & 38.91 \\
    EfficientPhys~\cite{efficientphys} & 1.91  & 373.72 & 8.42  & 26.68 \\
    PhysFormer~\cite{physformer} & 7.38  & 316.29 & 8.50  & 28.63 \\
    PhysFormer~(10s) & 7.38  & 316.29 & 10.27  & 22.17 \\
    Ours & 3.25  & 240.55 & 6.30  & 29.06 \\
    Ours~(10s) & 3.25  & 240.55  & 9.44  & 19.07 \\
    \bottomrule
    \end{tabular}%
  \caption{Computational cost.}
  \label{tab:cost}%
\end{table}%

\begin{figure*}[ht] 
  \centering  
  \includegraphics[width=0.99\linewidth]{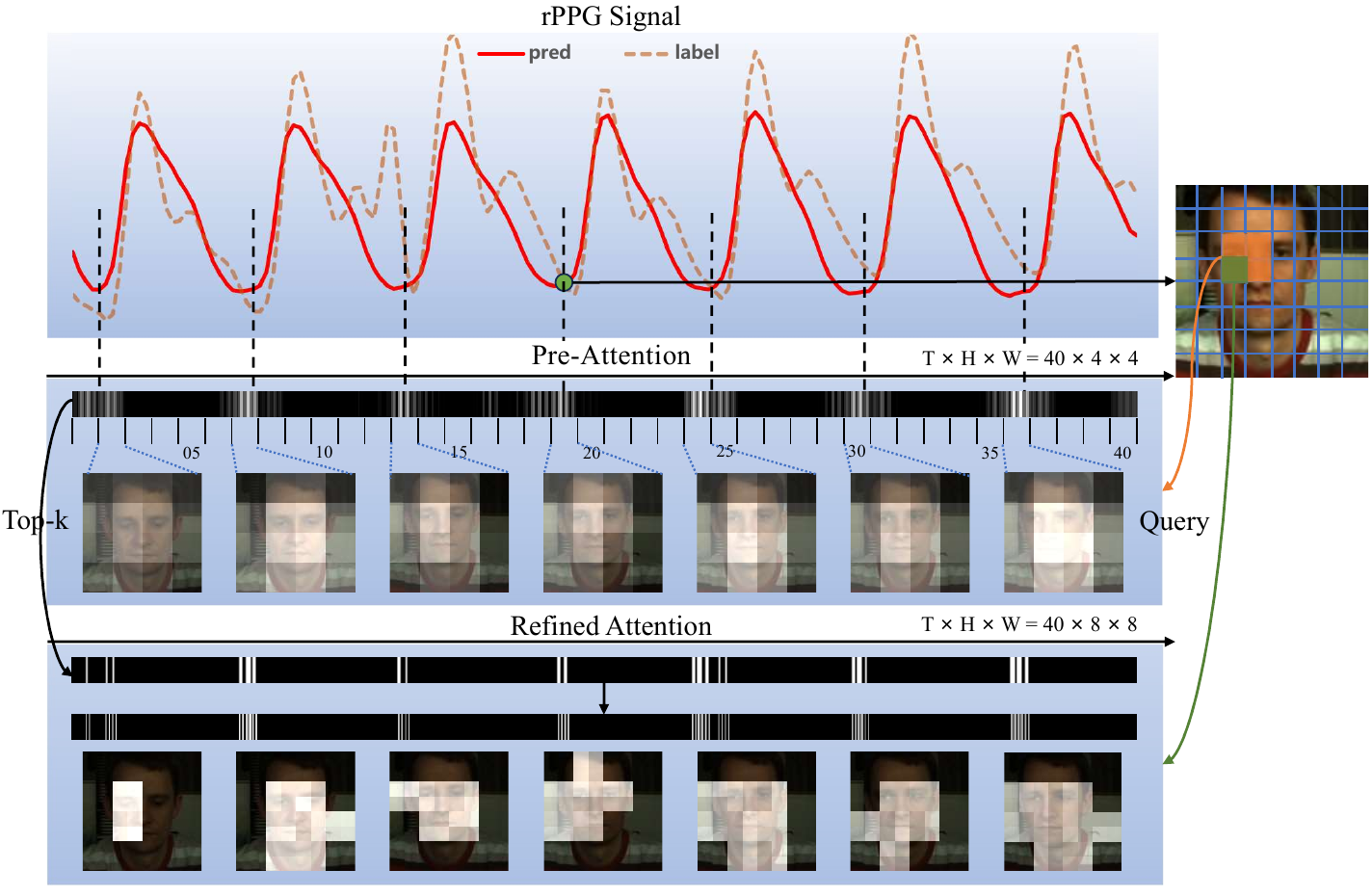}  
  \caption{Visualization of attention maps. Given the $294_{th}$ coarse-grained token in orange as a query for the pre-attention stage, while the $1179_{th}$ fine-grained token in green as a query for the refined attention stage. The refined attention is calculated in the top-k region obtained from the pre-attention. The figure displays attention score maps and key responses for both two attention stages.}
  \label{fig:vis}
\end{figure*}

\subsection{Visualization}
We visualized the periodic sparse attention map in Figure \ref{fig:vis}. For the pre-attention stage, given a token (the $294_{th}$) located at the temporal trough and the left face as a Query. The attention score map and key responses were obtained, where brighter regions indicate stronger attention. It provided the top-k (high-correlation) region for the refined attention stage to delve into rPPG features with finer granularity. For the refined attention stage, a sub-token (the $1179_{th}$) of the Query from the pre-attention was used as a Query in the top-k region. The attention score map and key responses were also obtained. As observed in Figure \ref{fig:vis}, the pre-attention score appropriately concentrates on skin regions with similar temporal phases, accurately expressing the periodic nature of rPPG. However, there are still many non-zero attention scores allocated to irrelevant areas. Refined attention focuses on the top-k regions, aiming to minimize attention on irrelevant areas as much as possible. This concentration of attention enables a more detailed extraction of rPPG features, reducing noise interference, and enhancing the model's robustness.

\subsection{Limitation}
We leverage the temporal attention sparsity induced by periodicity and introduce pre-attention to learn the periodic patterns. Although pre-attention operates at a coarse region level and does not generate significant computation, it inevitably leads to additional GPU kernel launches and memory transactions. 
From Table \ref{tab:cost}, RhythmFormer has lower computational complexity compared to Physformer, but its throughput is lower, likely due to this effect.
Nevertheless, this issue could be alleviated through engineering efforts, such as GPU kernel fusion.

%% file: sec/5_conclusion.tex
\label{sec/5_conclusion}
\section{Conclusion}

In response to the subtle and periodic nature of rPPG signals, we propose a method that explicitly leverages periodicity to design a periodic sparse attention mechanism. Combined with the proposed plug-and-play fusion stem to guide attention, this method effectively recognizes quasi-periodic patterns in weak rPPG signals, achieving state-of-the-art performance in extensive experiments. We envision RhythmFormer as a baseline for the rPPG community, supporting diverse methods and applications in the rPPG domain.
In the future, it is potential to explore multimodal periodic sparse attention, as it may align periodic patterns across various modalities, such as RGB, radar, and radio frequency. Additionally, periodic sparse attention has proven highly effective in the rPPG task, and we will investigate its feasibility for broader fine-grained or periodic video understanding tasks.

%% file: main.bbl
\begin{thebibliography}{10}
\expandafter\ifx\csname url\endcsname\relax
  \def\url#1{\texttt{#1}}\fi
\expandafter\ifx\csname urlprefix\endcsname\relax\def\urlprefix{URL }\fi
\expandafter\ifx\csname href\endcsname\relax
  \def\href#1#2{#2} \def\path#1{#1}\fi

\bibitem{GREEN}
W.~Verkruysse, L.~O. Svaasand, J.~S. Nelson, Remote plethysmographic imaging using ambient light., Optics express 16~(26) (2008) 21434--21445.

\bibitem{deepphys}
W.~Chen, D.~McDuff, Deepphys: Video-based physiological measurement using convolutional attention networks, in: Proceedings of the European Conference on Computer Vision (ECCV), 2018, pp. 349--365.

\bibitem{carrera2019pr}
D.~Carrera, B.~Rossi, P.~Fragneto, G.~Boracchi, Online anomaly detection for long-term ecg monitoring using wearable devices, Pattern Recognition 88 (2019) 482--492.

\bibitem{liu2022pr}
S.~Liu, J.~Han, E.~L. Puyal, S.~Kontaxis, S.~Sun, P.~Locatelli, J.~Dineley, F.~B. Pokorny, G.~Dalla~Costa, L.~Leocani, et~al., Fitbeat: Covid-19 estimation based on wristband heart rate using a contrastive convolutional auto-encoder, Pattern recognition 123 (2022) 108403.

\bibitem{pyvhr}
G.~Boccignone, D.~Conte, V.~Cuculo, A.~D’Amelio, G.~Grossi, R.~Lanzarotti, E.~Mortara, pyvhr: a python framework for remote photoplethysmography, PeerJ Computer Science 8 (2022) e929.

\bibitem{review1}
D.~McDuff, Camera measurement of physiological vital signs, ACM Computing Surveys 55~(9) (2023) 1--40.

\bibitem{apptac23}
C.~{\'A}. Casado, M.~L. Ca{\~n}ellas, M.~B. L{\'o}pez, Depression recognition using remote photoplethysmography from facial videos, IEEE Transactions on Affective Computing 14~(4) (2023) 3305--3316.

\bibitem{POS}
W.~Wang, A.~C. Den~Brinker, S.~Stuijk, G.~De~Haan, Algorithmic principles of remote ppg, IEEE Transactions on Biomedical Engineering 64~(7) (2016) 1479--1491.

\bibitem{casado2023face2ppg}
C.~A. Casado, M.~B. L{\'o}pez, Face2ppg: An unsupervised pipeline for blood volume pulse extraction from faces, IEEE Journal of Biomedical and Health Informatics (2023).

\bibitem{vivit}
A.~Arnab, M.~Dehghani, G.~Heigold, C.~Sun, M.~Lu{\v{c}}i{\'c}, C.~Schmid, Vivit: A video vision transformer, in: Proceedings of the IEEE/CVF International Conference on Computer Vision, 2021, pp. 6836--6846.

\bibitem{physformer}
Z.~Yu, Y.~Shen, J.~Shi, H.~Zhao, P.~H. Torr, G.~Zhao, Physformer: Facial video-based physiological measurement with temporal difference transformer, in: Proceedings of the IEEE/CVF Conference on Computer Vision and Pattern Recognition, 2022, pp. 4186--4196.

\bibitem{wolfe2019preattentive}
J.~M. Wolfe, I.~S. Utochkin, What is a preattentive feature?, Current opinion in psychology 29 (2019) 19--26.

\bibitem{wolfe2021guided}
J.~M. Wolfe, Guided search 6.0: An updated model of visual search, Psychonomic Bulletin \& Review 28~(4) (2021) 1060--1092.

\bibitem{CHROM}
G.~De~Haan, V.~Jeanne, Robust pulse rate from chrominance-based rppg, IEEE Transactions on Biomedical Engineering 60~(10) (2013) 2878--2886.

\bibitem{HR-CNN}
R.~{\v{S}}petl{\'\i}k, V.~Franc, J.~Matas, Visual heart rate estimation with convolutional neural network, in: Proceedings of the British Machine Vision Conference, Newcastle, UK, 2018, pp. 3--6.

\bibitem{rhythmNet}
X.~Niu, S.~Shan, H.~Han, X.~Chen, Rhythmnet: End-to-end heart rate estimation from face via spatial-temporal representation, IEEE Transactions on Image Processing 29 (2019) 2409--2423.

\bibitem{TSCAN}
X.~Liu, J.~Fromm, S.~Patel, D.~McDuff, Multi-task temporal shift attention networks for on-device contactless vitals measurement, Advances in Neural Information Processing Systems 33 (2020) 19400--19411.

\bibitem{zhang2025}
Y.~Zhang, H.~Lu, X.~Liu, Y.~Chen, K.~Wu, Advancing generalizable remote physiological measurement through the integration of explicit and implicit prior knowledge (2025).
\newblock \href {http://arxiv.org/abs/2403.06947} {\path{arXiv:2403.06947}}.

\bibitem{physnet}
Z.~Yu, X.~Li, G.~Zhao, Remote photoplethysmograph signal measurement from facial videos using spatio-temporal networks, in: 30th British Machine Visison Conference: BMVC 2019. 9th-12th September 2019, Cardiff, UK, The British Machine Vision Conference (BMVC), 2019.

\bibitem{3DCDC}
Y.~Zhao, B.~Zou, F.~Yang, L.~Lu, A.~N. Belkacem, C.~Chen, Video-based physiological measurement using 3d central difference convolution attention network, in: 2021 IEEE International Joint Conference on Biometrics (IJCB), IEEE, 2021, pp. 1--6.

\bibitem{contrastphys+}
Z.~Sun, X.~Li, Contrast-phys+: Unsupervised and weakly-supervised video-based remote physiological measurement via spatiotemporal contrast, IEEE Transactions on Pattern Analysis and Machine Intelligence 46~(8) (2024) 5835--5851.
\newblock \href {https://doi.org/10.1109/TPAMI.2024.3367910} {\path{doi:10.1109/TPAMI.2024.3367910}}.

\bibitem{yue2023}
Z.~Yue, M.~Shi, S.~Ding, Facial video-based remote physiological measurement via self-supervised learning, IEEE Transactions on Pattern Analysis and Machine Intelligence (2023).

\bibitem{efficientphys}
X.~Liu, B.~Hill, Z.~Jiang, S.~Patel, D.~McDuff, Efficientphys: Enabling simple, fast and accurate camera-based cardiac measurement, in: Proceedings of the IEEE/CVF Winter Conference on Applications of Computer Vision, 2023, pp. 5008--5017.

\bibitem{liu2024spikingphysformer}
M.~Liu, J.~Tang, H.~Li, J.~Qi, S.~Li, K.~Wang, Y.~Wang, H.~Chen, Spiking-physformer: Camera-based remote photoplethysmography with parallel spike-driven transformer (2024).
\newblock \href {http://arxiv.org/abs/2402.04798} {\path{arXiv:2402.04798}}.

\bibitem{transformer}
A.~Vaswani, N.~Shazeer, N.~Parmar, J.~Uszkoreit, L.~Jones, A.~N. Gomez, {\L}.~Kaiser, I.~Polosukhin, Attention is all you need, Advances in neural information processing systems 30 (2017).

\bibitem{swintransformer}
Z.~Liu, Y.~Lin, Y.~Cao, H.~Hu, Y.~Wei, Z.~Zhang, S.~Lin, B.~Guo, Swin transformer: Hierarchical vision transformer using shifted windows, in: Proceedings of the IEEE/CVF International Conference on Computer Vision, 2021, pp. 10012--10022.

\bibitem{mvit}
Z.~Tu, H.~Talebi, H.~Zhang, F.~Yang, P.~Milanfar, A.~Bovik, Y.~Li, Maxvit: Multi-axis vision transformer, in: European Conference on Computer Vision, Springer, 2022, pp. 459--479.

\bibitem{biformer}
L.~Zhu, X.~Wang, Z.~Ke, W.~Zhang, R.~W. Lau, Biformer: Vision transformer with bi-level routing attention, in: Proceedings of the IEEE/CVF Conference on Computer Vision and Pattern Recognition, 2023, pp. 10323--10333.

\bibitem{LCE}
S.~Ren, D.~Zhou, S.~He, J.~Feng, X.~Wang, Shunted self-attention via multi-scale token aggregation, in: Proceedings of the IEEE/CVF Conference on Computer Vision and Pattern Recognition, 2022, pp. 10853--10862.

\bibitem{dualgan}
H.~Lu, H.~Han, S.~K. Zhou, Dual-gan: Joint bvp and noise modeling for remote physiological measurement, in: Proceedings of the IEEE/CVF Conference on Computer Vision and Pattern Recognition, 2021, pp. 12404--12413.

\bibitem{rppgmae}
X.~Liu, Y.~Zhang, Z.~Yu, H.~Lu, H.~Yue, J.~Yang, rppg-mae: Self-supervised pretraining with masked autoencoders for remote physiological measurements, IEEE Transactions on Multimedia (2024).

\bibitem{physformer++}
Z.~Yu, Y.~Shen, J.~Shi, H.~Zhao, Y.~Cui, J.~Zhang, P.~Torr, G.~Zhao, Physformer++: Facial video-based physiological measurement with slowfast temporal difference transformer, International Journal of Computer Vision 131~(6) (2023) 1307--1330.

\bibitem{tdn}
L.~Wang, Z.~Tong, B.~Ji, G.~Wu, Tdn: Temporal difference networks for efficient action recognition, in: 2021 IEEE/CVF Conference on Computer Vision and Pattern Recognition (CVPR), 2021, pp. 1895--1904.
\newblock \href {https://doi.org/10.1109/CVPR46437.2021.00193} {\path{doi:10.1109/CVPR46437.2021.00193}}.

\bibitem{mul1}
H.~Hu, S.~Dong, Y.~Zhao, D.~Lian, Z.~Li, S.~Gao, Transrac: Encoding multi-scale temporal correlation with transformers for repetitive action counting, in: Proceedings of the IEEE/CVF Conference on Computer Vision and Pattern Recognition, 2022, pp. 19013--19022.

\bibitem{mul3}
R.~Dai, S.~Das, K.~Kahatapitiya, M.~S. Ryoo, F.~Br{\'e}mond, Ms-tct: multi-scale temporal convtransformer for action detection, in: Proceedings of the IEEE/CVF Conference on Computer Vision and Pattern Recognition, 2022, pp. 20041--20051.

\bibitem{TDC}
Z.~Yu, B.~Zhou, J.~Wan, P.~Wang, H.~Chen, X.~Liu, S.~Z. Li, G.~Zhao, Searching multi-rate and multi-modal temporal enhanced networks for gesture recognition, IEEE Transactions on Image Processing 30 (2021) 5626--5640.

\bibitem{autohr}
Z.~Yu, X.~Li, X.~Niu, J.~Shi, G.~Zhao, Autohr: A strong end-to-end baseline for remote heart rate measurement with neural searching, IEEE Signal Processing Letters 27 (2020) 1245--1249.

\bibitem{PURE}
R.~Stricker, S.~M{\"u}ller, H.-M. Gross, Non-contact video-based pulse rate measurement on a mobile service robot, in: The 23rd IEEE International Symposium on Robot and Human Interactive Communication, IEEE, 2014, pp. 1056--1062.

\bibitem{UBFC-rPPG}
S.~Bobbia, R.~Macwan, Y.~Benezeth, A.~Mansouri, J.~Dubois, Unsupervised skin tissue segmentation for remote photoplethysmography, Pattern Recognition Letters 124 (2019) 82--90.

\bibitem{cohface}
G.~Heusch, A.~Anjos, S.~Marcel, A reproducible study on remote heart rate measurement, arXiv preprint arXiv:1709.00962 (2017).

\bibitem{MMPD}
J.~Tang, K.~Chen, Y.~Wang, Y.~Shi, S.~Patel, D.~McDuff, X.~Liu, Mmpd: Multi-domain mobile video physiology dataset, in: 45th Annual International Conference of the IEEE Engineering in Medicine and Biology Society, 2023.

\bibitem{toolbox}
X.~Liu, G.~Narayanswamy, A.~Paruchuri, X.~Zhang, J.~Tang, Y.~Zhang, S.~Sengupta, S.~Patel, Y.~Wang, D.~McDuff, \href{https://openreview.net/forum?id=q4XNX15kSe}{r{PPG}-toolbox: Deep remote {PPG} toolbox}, in: Thirty-seventh Conference on Neural Information Processing Systems Datasets and Benchmarks Track, 2023.
\newline\urlprefix\url{https://openreview.net/forum?id=q4XNX15kSe}

\bibitem{pulsegan}
R.~Song, H.~Chen, J.~Cheng, C.~Li, Y.~Liu, X.~Chen, Pulsegan: Learning to generate realistic pulse waveforms in remote photoplethysmography, IEEE Journal of Biomedical and Health Informatics 25~(5) (2021) 1373--1384.

\bibitem{synrhythm}
X.~Niu, H.~Han, S.~Shan, X.~Chen, Synrhythm: Learning a deep heart rate estimator from general to specific, in: 2018 24th International Conference on Pattern Recognition (ICPR), IEEE, 2018, pp. 3580--3585.

\bibitem{Meta-rppg}
E.~Lee, E.~Chen, C.-Y. Lee, Meta-rppg: Remote heart rate estimation using a transductive meta-learner, in: Computer Vision--ECCV 2020: 16th European Conference, Glasgow, UK, August 23--28, 2020, Proceedings, Part XXVII 16, Springer, 2020, pp. 392--409.

\bibitem{CVD}
X.~Niu, Z.~Yu, H.~Han, X.~Li, S.~Shan, G.~Zhao, Video-based remote physiological measurement via cross-verified feature disentangling, in: Computer Vision--ECCV 2020: 16th European Conference, Glasgow, UK, August 23--28, 2020, Proceedings, Part II 16, Springer, 2020, pp. 295--310.

\bibitem{deeprppg}
S.-Q. Liu, P.~C. Yuen, A general remote photoplethysmography estimator with spatiotemporal convolutional network, in: 2020 15th IEEE International Conference on Automatic Face and Gesture Recognition (FG 2020), 2020, pp. 481--488.
\newblock \href {https://doi.org/10.1109/FG47880.2020.00109} {\path{doi:10.1109/FG47880.2020.00109}}.

\bibitem{AND-rPPG}
B.~Lokendra, G.~Puneet, And-rppg: A novel denoising-rppg network for improving remote heart rate estimation, Computers in biology and medicine 141 (2022) 105146.

\bibitem{etarppgnet}
M.~Hu, F.~Qian, D.~Guo, X.~Wang, L.~He, F.~Ren, Eta-rppgnet: Effective time-domain attention network for remote heart rate measurement, IEEE Transactions on Instrumentation and Measurement 70 (2021) 1--12.
\newblock \href {https://doi.org/10.1109/TIM.2021.3058983} {\path{doi:10.1109/TIM.2021.3058983}}.

\bibitem{AAAI23}
J.~Li, Z.~Yu, J.~Shi, Learning motion-robust remote photoplethysmography through arbitrary resolution videos, in: Proceedings of the AAAI Conference on Artificial Intelligence, Vol.~37, 2023, pp. 1334--1342.

\bibitem{sinc}
J.~Speth, N.~Vance, P.~Flynn, A.~Czajka, Non-contrastive unsupervised learning of physiological signals from video, in: Proceedings of the IEEE/CVF Conference on Computer Vision and Pattern Recognition, 2023, pp. 14464--14474.

\bibitem{iccv23}
Z.~Li, L.~Yin, Contactless pulse estimation leveraging pseudo labels and self-supervision, in: 2023 IEEE/CVF International Conference on Computer Vision (ICCV), 2023, pp. 20531--20540.
\newblock \href {https://doi.org/10.1109/ICCV51070.2023.01882} {\path{doi:10.1109/ICCV51070.2023.01882}}.

\bibitem{ICA}
M.-Z. Poh, D.~J. McDuff, R.~W. Picard, Non-contact, automated cardiac pulse measurements using video imaging and blind source separation., Optics express 18~(10) (2010) 10762--10774.

\bibitem{LGI}
C.~S. Pilz, S.~Zaunseder, J.~Krajewski, V.~Blazek, Local group invariance for heart rate estimation from face videos in the wild, in: Proceedings of the IEEE Conference on Computer Vision and Pattern Recognition Workshops, 2018, pp. 1254--1262.

\bibitem{PBV}
G.~De~Haan, A.~Van~Leest, Improved motion robustness of remote-ppg by using the blood volume pulse signature, Physiological measurement 35~(9) (2014) 1913.

\end{thebibliography}
